\documentclass{article}

\usepackage{arxiv}
\usepackage[numbers]{natbib} 
\usepackage[utf8]{inputenc} 
\usepackage[T1]{fontenc}    
\usepackage{hyperref}       
\usepackage{url}            
\usepackage{booktabs}       
\usepackage{amsfonts}       
\usepackage{nicefrac}       
\usepackage{microtype}      
\usepackage{lipsum}
\usepackage{graphicx}
\graphicspath{ {./images/} }

\usepackage{graphicx}%
\usepackage{multirow}%
\usepackage{amsmath,amssymb,amsfonts}%
\usepackage{amsthm}%
\usepackage{mathrsfs}%
\usepackage[title]{appendix}%
\usepackage{xcolor}%
\usepackage{textcomp}%
\usepackage{manyfoot}%
\usepackage{booktabs}%
\usepackage{algorithm}%
\usepackage{algorithmicx}%
\usepackage{algpseudocode}%
\usepackage{listings}%
\usepackage{subfig}
\usepackage{caption}
\usepackage{comment}

\usepackage[utf8]{inputenc}     
\usepackage[T1]{fontenc}        
\usepackage{nicefrac}           
\usepackage{microtype}          
\usepackage{bm}                 
\usepackage{enumitem}           
\usepackage{float}              
\usepackage[export]{adjustbox}  
\usepackage{amsmath}
\usepackage{colortbl}

\newcommand{\resultnof}[2]{#1$\pm$\footnotesize{#2}}

\newcommand{\edit}[1]{{\color{black}#1}}

\usepackage[utf8]{inputenc}
\usepackage{soul}
\usepackage{pifont}
\usepackage{amsmath}
\usepackage{amssymb}
\usepackage{bm}
\usepackage{color,soul}
\usepackage[dvipsnames]{xcolor}
\usepackage{gensymb}

\usepackage{xcolor}

\def\btheta{\bm{\theta}}

\newcommand{\mytilde}{\raise.17ex\hbox{$\scriptstyle\mathtt{\sim}$}}

\let\emptyset\varnothing

\def\pp{\mathbf{p}}

\def\xx{\mathbf{x}}

\def\zz{\mathbf{z}}

\def\II{\mathbf{I}}

\def\bB{\mathcal{B}}
\def\cC{\mathcal{C}}
\def\dD{\mathcal{D}}

\def\iI{\mathcal{I}}

\def\lL{\mathcal{L}}

\def\nN{\mathcal{N}}

\def\pP{\mathcal{P}}

\def\Ee{\mathbb{E}}

\def\Re{\mathbb{R}}

\usepackage{amsmath,amsfonts,bm}

\def\eqref#1{equation~\ref{#1}}

\def\1{\bm{1}}

\DeclareMathAlphabet{\mathsfit}{\encodingdefault}{\sfdefault}{m}{sl}
\SetMathAlphabet{\mathsfit}{bold}{\encodingdefault}{\sfdefault}{bx}{n}

\DeclareMathOperator*{\argmin}{arg\,min}

\renewcommand{\today}{}
\makeatletter
\date{}
\def\@date{}
\makeatother

\title{\textit{Dream2Learn}: Structured Generative Dreaming for Continual Learning}

\author{
  Salvatore Calcagno$^1$, Matteo Pennisi$^1$, Federica Proietto Salanitri$^1$, \\ 
  \textbf{Amelia Sorrenti$^1$, Simone Palazzo$^1$, Concetto Spampinato$^1$, Giovanni Bellitto$^1$} \\
  $^1$PeRCeiVe Lab, University of Catania, Italy \\
  \texttt{\{salvatore.calcagno, amelia.sorrenti\}@phd.unict.it} \\
  \texttt{\{matteo.pennisi, federica.proiettosalanitri, simone.palazzo\}@unict.it} \\
  \texttt{\{concetto.spampinato, giovanni.bellitto\}@unict.it}
}

\begin{document}
\maketitle

\begin{figure*}[htb!]
    \centering
    \setlength{\tabcolsep}{1pt}
    \renewcommand{\arraystretch}{1}
    \resizebox{0.8\textwidth}{!}{
    \begin{tabular}{ccccccc}
        \rotatebox{90}{~~~Scoreboard} &
        \subfloat{\includegraphics[width=0.15\textwidth]{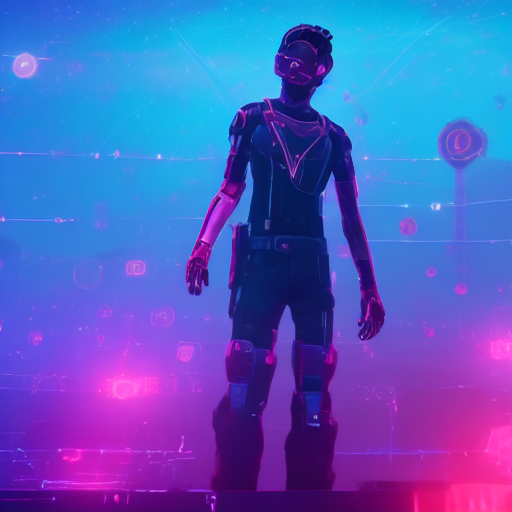}} &
        \subfloat{\includegraphics[width=0.15\textwidth]{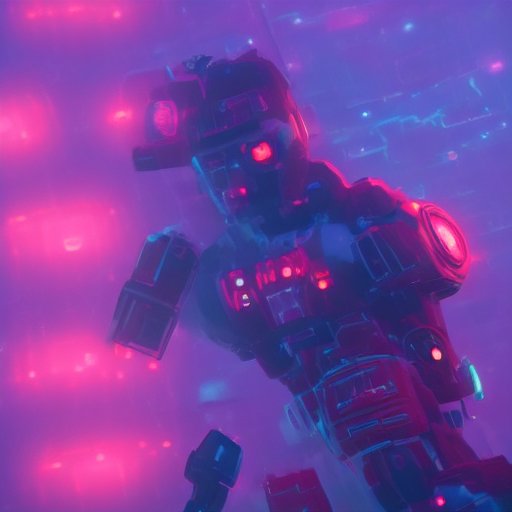}} &
        \subfloat{\includegraphics[width=0.15\textwidth]{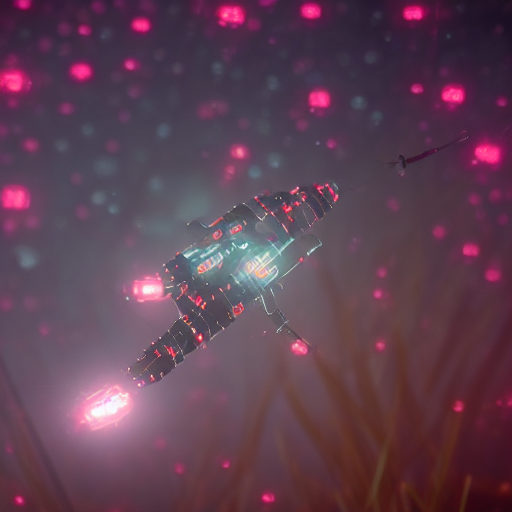}} &
        \subfloat{\includegraphics[width=0.15\textwidth]{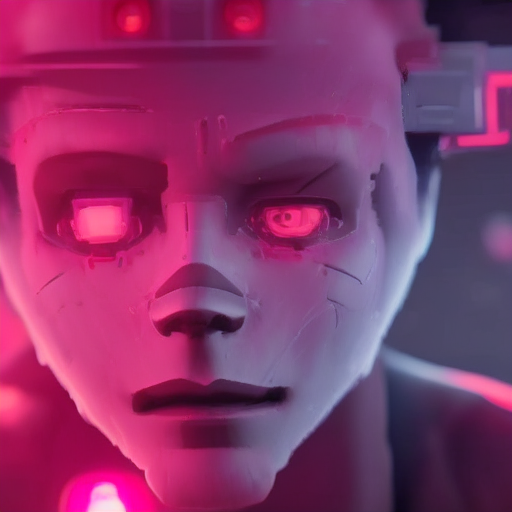}} &
        \subfloat{\includegraphics[width=0.15\textwidth]{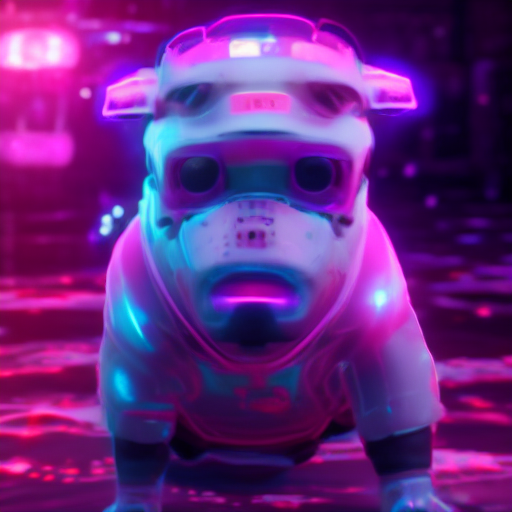}} &
        \subfloat{\includegraphics[width=0.15\textwidth]{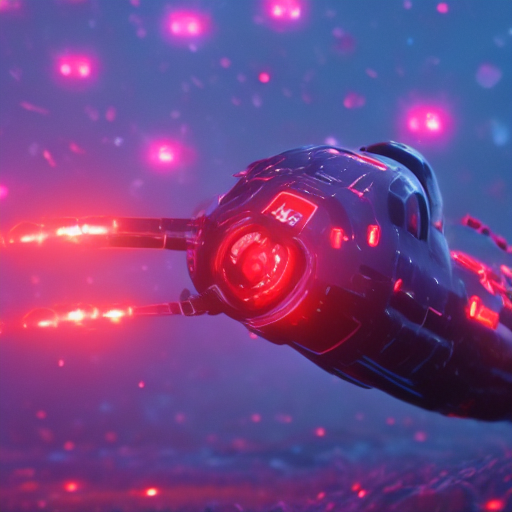}} \\

        \rotatebox{90}{~~~~~Ladybug} &
        \subfloat{\includegraphics[width=0.15\textwidth]{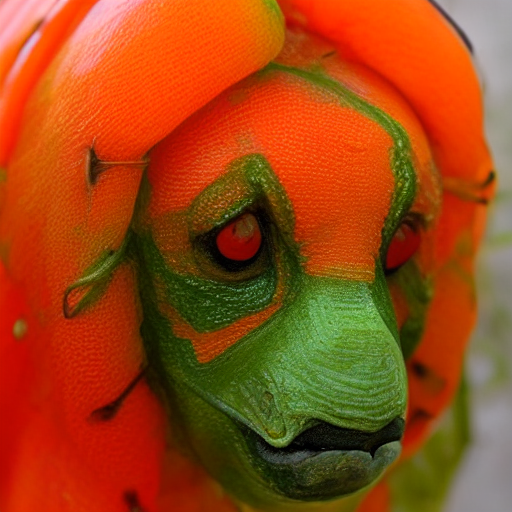}} &
        \subfloat{\includegraphics[width=0.15\textwidth]{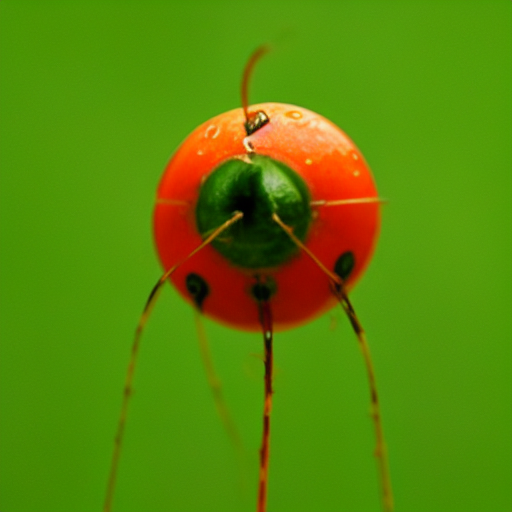}} &
        \subfloat{\includegraphics[width=0.15\textwidth]{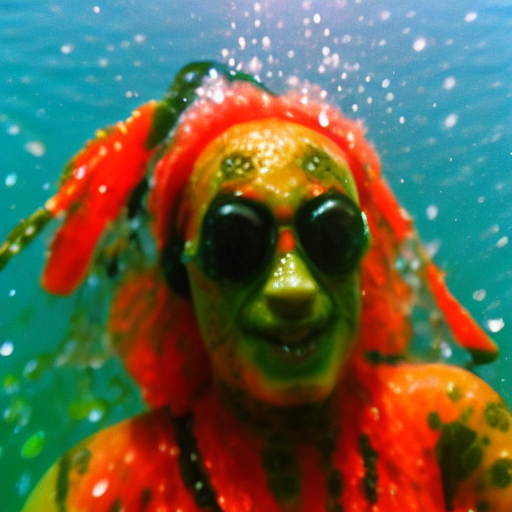}} &
        \subfloat{\includegraphics[width=0.15\textwidth]{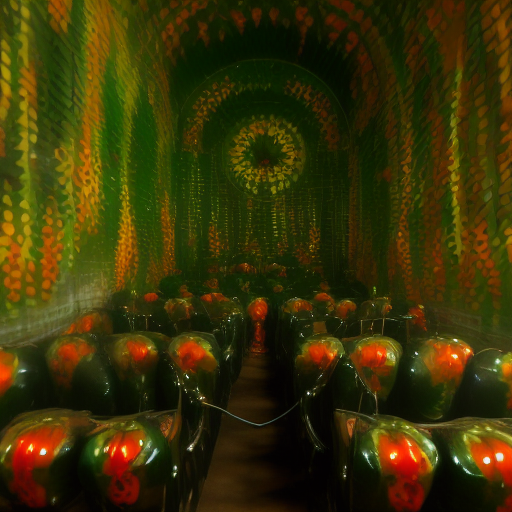}} &
        \subfloat{\includegraphics[width=0.15\textwidth]{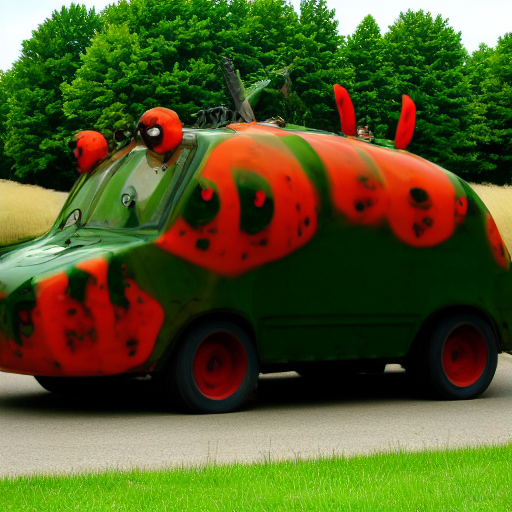}} &
        \subfloat{\includegraphics[width=0.15\textwidth]{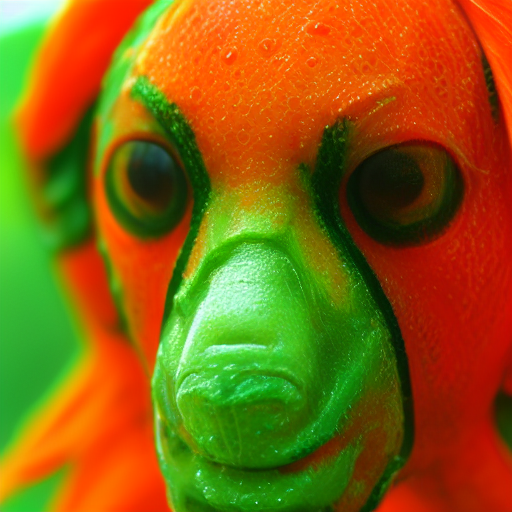}} \\
        
        \rotatebox{90}{~~~~School bus} &
        \subfloat{\includegraphics[width=0.15\textwidth]{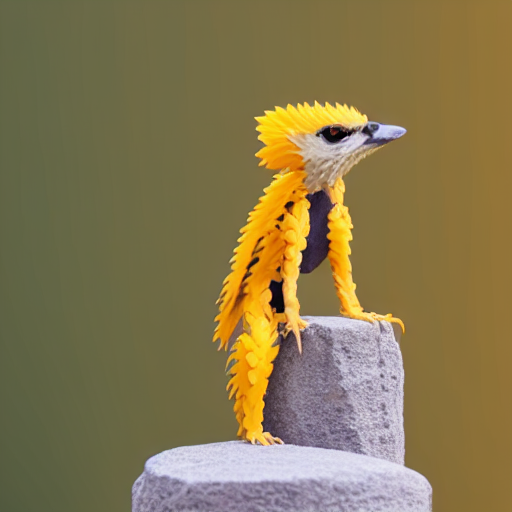}} &
        \subfloat{\includegraphics[width=0.15\textwidth]{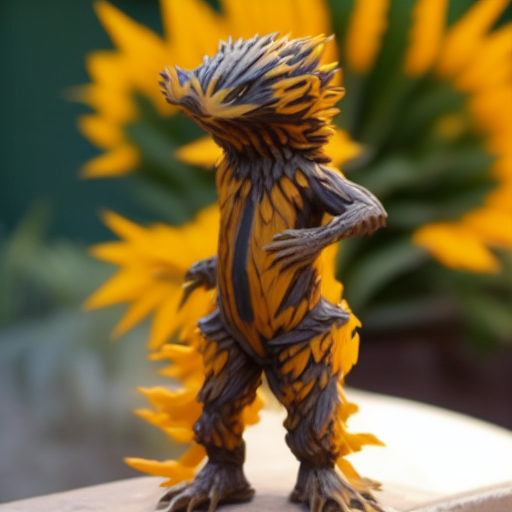}} &
        \subfloat{\includegraphics[width=0.15\textwidth]{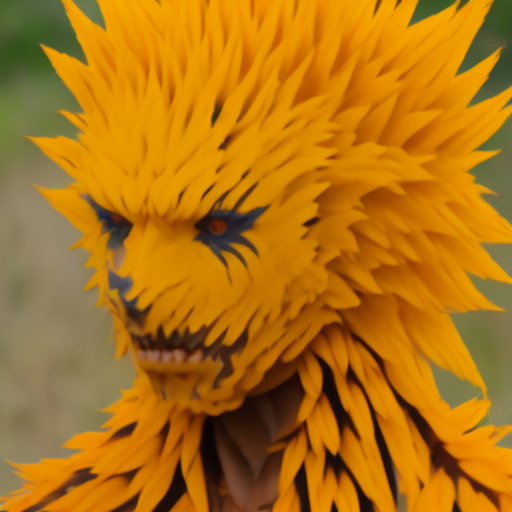}} &
        \subfloat{\includegraphics[width=0.15\textwidth]{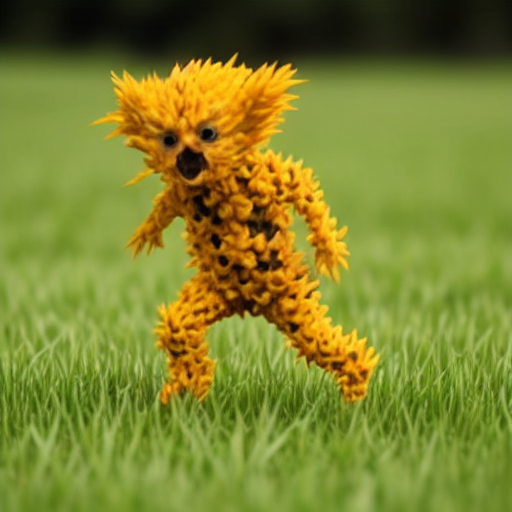}} &
        \subfloat{\includegraphics[width=0.15\textwidth]{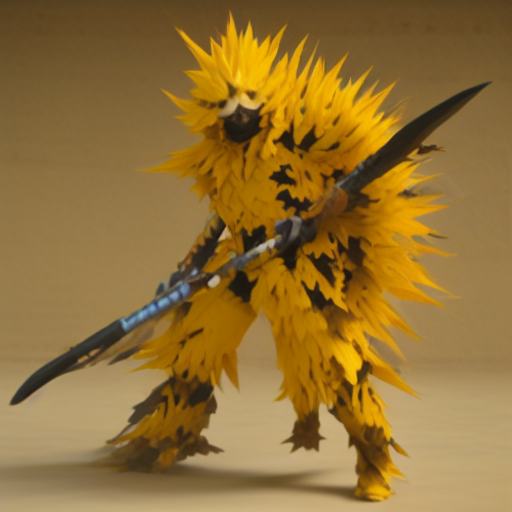}} &
        \subfloat{\includegraphics[width=0.15\textwidth]{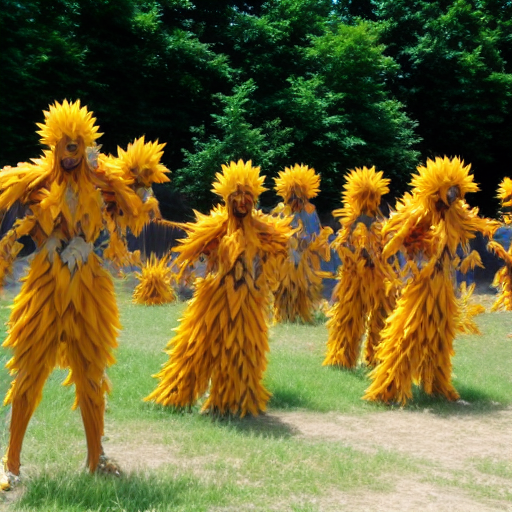}} \\
    \end{tabular}
    }
    \caption{\textbf{Dreamed classes generated by D2L}. Examples of dreamed classes synthesized from their corresponding real classes (left). These samples emerge as semantically distinct yet \textbf{structurally coherent representations} in the generator's latent space, forming intermediate concepts that enhance the continual classifier's generalization to future tasks.}
    \label{fig:teaser}
    
\end{figure*}

\begin{abstract}
Continual learning requires balancing plasticity and stability while mitigating catastrophic forgetting. Inspired by human dreaming as a mechanism for internal simulation and knowledge restructuring, we introduce \textbf{Dream2Learn (D2L)}, a framework in which a model autonomously generates structured synthetic experiences from its own internal representations and uses them for self-improvement. Rather than reconstructing past data as in generative replay, D2L enables a classifier to create novel, semantically distinct dreamed classes that are coherent with its learned knowledge yet do not correspond to previously observed data. These dreamed samples are produced by conditioning a frozen diffusion model through soft prompt optimization driven by the classifier itself. The generated data are not used to replace memory, but to expand and reorganize the representation space, effectively allowing the network to self-train on internally synthesized concepts. By integrating dreamed classes into continual training, D2L proactively structures latent features to support forward knowledge transfer and adaptation to future tasks. This prospective self-training mechanism mirrors the role of sleep in consolidating and reorganizing memory, turning internal simulations into a tool for improved generalization. Experiments on Mini-ImageNet, FG-ImageNet, and ImageNet-R demonstrate that D2L consistently outperforms strong rehearsal-based baselines and achieves positive forward transfer, confirming its ability to enhance adaptability through internally generated training signals.

\end{abstract}


\section{Introduction}
Humans possess a remarkable ability to learn continuously, consolidate past experiences, and generalize knowledge to novel situations~\citep{cls_2026, cls_1995}. This process is also facilitated by memory replay and restructuring during sleep, where the brain synthesizes realistic dreams derived from awake experiences to prepare for future challenges~\citep{pmid17173043, pmid15450165, shapiro_pnas}. 
In contrast, deep learning models in continual learning (CL) suffer from catastrophic forgetting, wherein previously acquired knowledge deteriorates when new tasks are introduced~\citep{mccloskey1989catastrophic}. 
Traditional CL methods attempt to address this issue through rehearsal-based strategies, regularization techniques, or architectural modifications. However, they often struggle to effectively balance stability and plasticity, thereby limiting both long-term knowledge retention and the capacity for adaptation~\citep{De_Lange_2021_ICCV, parisi2019continual}.\
Among these, rehearsal-based strategies are widely used due to their ability to stabilize learning by replaying stored examples. Yet, despite their effectiveness, such approaches diverge significantly from how the human brain consolidates memory. Rather than relying on the exact replay of past experiences, the brain engages in generative processes during dreaming, recombining perceptual elements from daily life to construct novel and plausible future scenarios~\citep{pmid26779078, Schwartz2003}.
This process allows for efficient knowledge reinforcement, enabling the brain to improve generalization and anticipate future challenges. Translating this process into artificial neural networks is non-trivial, as it requires the ability to synthesize meaningful and structured representations of the previously learned knowledge without relying on external supervision.\\
To accomplish this task, in this paper we propose \textbf{Dream2Learn (D2L)}, a generative dreaming process that synthesizes training samples directly from the classifier’s internal representations. Unlike WSCL~\citep{sorrenti2023}, which relies on surrogate real data to simulate dreams, and other sleep-based approaches that primarily reinforce existing representations~\citep{tadros2022sleep, harun2023siesta}, D2L autonomously constructs future-adaptive representations (the \textit{dreams}, indeed), ensuring task relevance and enhancing the model’s ability to generalize to new tasks. \\
As illustrated in Fig.~\ref{fig:teaser}, D2L generates structured dreamed classes that serve as intermediate representations, facilitating continual learning. Instead of merely blending past class features, these dreamed samples form coherent yet distinct new concepts, expanding the representation space in a way that supports future task adaptation. By integrating ``dreamed classes" into training, the classifier learns high-level reusable features, reinforcing forward transfer while mitigating catastrophic forgetting. This process mirrors the role of REM sleep, where synthetic experiences help refine learned representations, maintaining long-term adaptability as new data distributions emerge.\\
Our experiments on Mini-ImageNet, FG-ImageNet, and ImageNet-R show that our strategy significantly boosts performance when integrated with standard continual learning methods. 
\section{Related Work}

Continual Learning (CL)~\citep{de2019continual,parisi2019continual} encompasses a family of machine learning techniques that aim to develop models that learn incrementally while avoiding catastrophic forgetting~\citep{mccloskey1989catastrophic}. Common strategies include regularization techniques~\citep{kirkpatrick2017overcoming, zenke2017continual}, architectural modifications~\citep{schwarz2018progress, mallya2018packnet}, and replay-based methods~\citep{robins1995catastrophic, rebuffi2017icarl, buzzega2020dark}. More recent approaches enhance model robustness through contrastive learning~\citep{mai2021supervised, cha2021co2l} and latent space regularization~\citep{frascaroli2023casper}, while experience replay optimizes sample selection for efficient memory retention~\citep{aljundi2019gradient, chaudhry2021using}. 

Generative Replay (GR)~\citep{shin2017continual, rios2019closed, liu2020generative} has emerged as an alternative to buffer-based experience replay by synthesizing past samples, but early methods often faced mode collapse and underperformed compared to traditional replay. 
Although DDGR~\citep{gao2023ddgr}, SDDR~\citep{jodelet2023class}, and DiffClass~\citep{meng2024diffclass} improve sample fidelity, the role of GR remains retrospective: the generator mainly acts as a memory proxy, reconstructing prior distributions for rehearsal rather than proactively reorganizing representations.
Inspired by cognitive neuroscience, several works explore memory mechanisms modeled on brain function. DualNet~\citep{pham2021dualnet} and DualPrompt~\citep{wang2022dualprompt} introduce parallel learning pathways, while CLS-ER~\citep{arani2022learning} and FearNet~\citep{kemker2018fearnet} implement short- and long-term memory systems. These approaches focus on stabilizing representations during learning but do not incorporate offline processes for restructuring knowledge. Sleep-based learning offers a complementary perspective, drawing from evidence that wake-sleep cycles refine memory representations~\citep{hinton1995wake, pmid35384841}. Sleep Replay Consolidation~\citep{tadros2022sleep} applies Hebbian plasticity, and SIESTA~\citep{harun2023siesta} introduces intermittent consolidation to support online learning. WSCL~\citep{sorrenti2023} alternates wake and sleep cycles to simulate the benefits of dreaming for memory consolidation. However, instead of generating internal experiences during sleep phases, it shapes the latent space of the classifier using pre-defined representations, limiting the biological plausibility of the dreaming process.

D2L reframes generation as a prospective mechanism: rather than replicating past data for rehearsal, generation is used to proactively structure the representation space towards upcoming tasks. It introduces a self-sufficient generative dreaming mechanism, by generating additional training signals from the classifier’s internal representations, ensuring task relevance and autonomous dreaming. Through soft prompt optimization, it identifies semantically distinct yet structurally coherent classes in the diffusion model's latent space, which act as intermediate anchors that prime future learning dynamics.
Thus, unlike GR techniques that focus on reconstructing or augmenting past distributions~\citep{jodelet2023class,meng2024diffclass}, D2L actively shapes future-adaptive representations, transforming dreaming into a mechanism for fostering forward knowledge transfer and long-term retention, enabling the classifier to adapt more effectively to unseen tasks.
\begin{figure*}[t]
\includegraphics[width=1\textwidth]{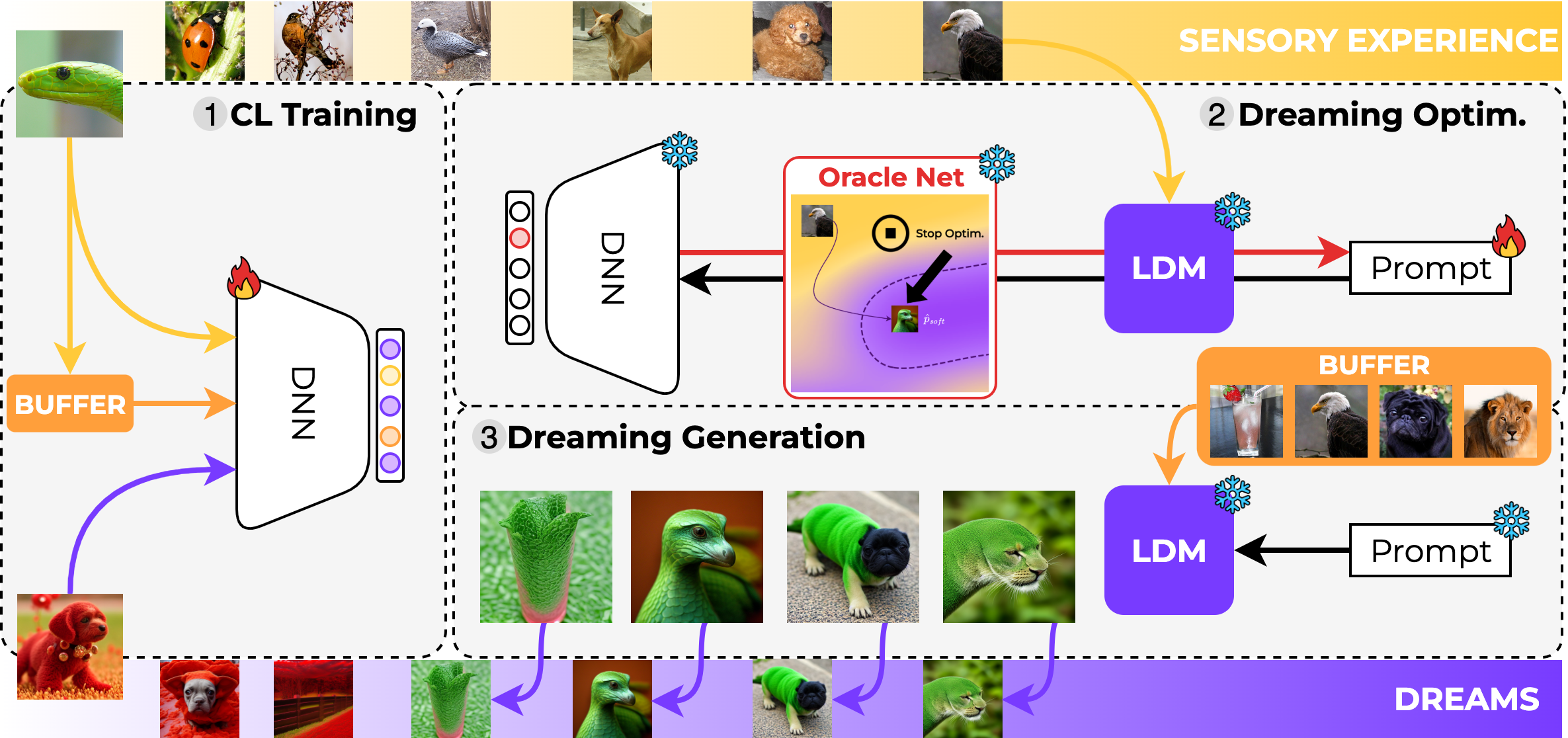}
\centering

\caption{\textbf{Overview of Dream2Learn.} (1) During CL training, a deep neural network (DNN) learns from real sensory images (the current task distribution plus the buffer) and dreamed samples produced by a latent diffusion model (LDM). (2) The dreaming optimization process refines the LDM prompts, with an Oracle Network providing a stopping criterion that prevents collapse. (3) Prompts generate auxiliary classes: dreamed samples are not buffered, but rather enrich the representation space with coherent latent clusters that foster knowledge reuse and adaptation.
}
\label{fig:method}
\vspace{-0.1cm}
\end{figure*}

\section{Method} 
\label{sec:method}

We formulate our continual learning setting as the problem of training a model $F_{\btheta}$ over a sequence of $T$ visual classification tasks $\left\{ \tau_1, \dots, \tau_T \right\}$, with each task $\tau_t$ associated to a dataset $\dD_t = \left\{ \left( \xx_{t,1}, y_{t,1} \right), \dots, \left( \xx_{t,{n_t}}, y_{t,{n_t}}  \right) \right\}$. Observations $\xx_{t,i}$ belong to an image space $\iI$, and class sets are disjoint across tasks, i.e., $y_{t,i} \in \cC_t$, $\cC_j \cap \cC_k = \emptyset$. The model's output layer has as many neurons as the total number of classes, i.e., $\sum n_t$. Notation-wise, we will treat $\dD_t$ as a probability distribution, when clear from the context.

We also assume the availability of a replay buffer $\bB$, where we store a limited number of samples from past tasks for rehearsal, and of a pre-trained and frozen image generator $G$. Our approach requires that $G$ can be conditioned from both textual prompts (with the possibility of adding learnable tokens) and input images; these requirements are easily satisfied by standard text-conditioned latent diffusion models (LDM) --- e.g., Stable Diffusion conditioned by CLIP embeddings~\citep{ramesh2022hierarchical, luo2023lcm} --- with an image adapter~\citep{ye2023ip}\footnote{We use \href{https://huggingface.co/h94/IP-Adapter}{h94/IP-Adapter}}. Formally, $G: \iI \times \pP \rightarrow \iI$, with $\pP$ being the space of sequences of textual token embeddings\footnote{In practice, $G$ is also made stochastic by receiving a random noise $\epsilon \sim \nN(\mathbf{0}, \II)$, which is omitted for brevity.}. We employ $G$, with appropriate conditioning, to synthesize dream images from past knowledge, thus creating an auxiliary synthetic data stream for preparation to future tasks. 
At the beginning of our procedure, the model $F_{\btheta}$ is trained to learn how to perform task $\tau_1$. Since no knowledge is initially present (as $\btheta$ is randomly initialized), we bootstrap the model by training it on task data $\dD_1$, optimizing a cross-entropy loss:
\begin{equation}
\min_{\btheta} \mathcal{L}_\text{CE}(F_{\btheta}, \dD_1) = -\mathbb{E}_{(\xx, y) \sim \dD_1} \Bigl[ \log p(y | \xx; \btheta) \Bigr] ,
\label{eq:ce_t1}
\end{equation}
with $p(y | \xx; \btheta)$ being the likelihood of the correct class, given model parameters $\btheta$. During this bootstrap phase, we also populate the buffer $\bB$ via reservoir sampling~\citep{lopez2017gradient}.

The bootstrapped classifier can now be used to synthesize new classes as ``dream'' variants of what the model has seen up to this point. Dream classes are created by optimizing a learnable prompt $\pp_c$ for each class $c\in \cC_1$, such that $G(\xx, \pp_c)$ transforms an input image $\xx$ into a ``dreamed'' versions that vaguely resemble target class $c$, thereby creating a synthetic distribution for a novel ``dream class''.
The details about this optimization process are given in Sec.~\ref{sec:dream}. 
Using this procedure, at task $\tau_1$ we introduce $n_1$ dream classes, i.e., as many as the current task's actual classes. We indicate with $\dD_1^d$ the distribution of dream classes defined at this stage. \\
Let \(\dD_{\tau_1}\) be the mixture distribution which equally samples from real data $\dD_1$ and from the dream distribution $\dD_1^d$. The classifier \(F_{\btheta}\) is then fine-tuned on \(\dD_{\tau_1}\), replacing \(\dD_1\) in Eq.~\ref{eq:ce_t1}. \\

On subsequent tasks $\tau_t$, $t>1$, we can exploit the model's knowledge on dream classes to ease its learning of new classes. At the beginning of $\tau_t$, we forward samples from each new class $c \in \cC_t$ through the model, and map $c$ to the output dream neuron with the largest average likelihood, as in~\citep{bellitto2022effects}. This allows to bootstrap each class maximizing the reuse of relevant features and preventing disrupting weight updates (details on this procedure in Sec.~\ref{sec:dream}). 
The dream classes corresponding to the assigned classification heads are removed.
We then train $F_{\btheta}$ on task data $\dD_t$ and on \edit{ $\dD_{t-1}^{d*}$, the residual dream distribution obtained from $\dD_{t-1}^{d}$ by removing the discarded dream classes $\dD_{t}^r$}, optimizing the following:
\begin{equation}
\min_{\btheta} \Bigl[ 
\lL_{\text{CE}}(F_{\btheta}, \dD_{t-1}^{d*} \cup \dD_t) + 
\lL_{\text{CL}}(F_{\btheta}, \dD_t, \bB) 
\Bigr] ,
\label{eq:ce_tn}
\end{equation}
where \edit{$\lL_{\text{CL}}$ is an additional continual learning loss that counters forgetting and explicitly leverages the replay buffer $\bB$ for rehearsal.}
In practice, when sampling from the dream distributions $\dD_{t-1}^{d*}$, we employ items stored in the buffer as input conditions to the generator $G$, to increase variability in the dreamed images. \edit{ Importantly, dreamed samples are never added to $\bB$ only their prompts are retained as part of a persistent dream inventory.}
After training on task $\tau_t$ and storing rehearsal samples into $\bB$, we update the dream inventory for the next task, by optimizing a new set of prompts $\left\{\pp_c ~|~ c \in \cC_t \right\}$,
corresponding to new dream class distributions. The set of $n_t$ newly-generated dream distributions is used to replace \edit{an equal number of existing} dreaming classes using again the mapping strategy in~\citep{bellitto2022effects}. 
An overview of the method is depicted in Fig.~\ref{fig:method}.

\subsection{Dreaming optimization and mapping}
\label{sec:dream}
The dreaming optimization process for task \( \tau_t \) consists of learning a proper conditioning for generator $G$, in order to synthesize samples of novel concepts, expanding the model’s representation space while preserving feature reuse.\\

For each real task class \( c \in \cC_t \), we aim to generate a corresponding dreamed class \( c^d \) that is distinct from all real classes, while contributing to a structured representation in the latent space. \\
To achieve this, we optimize a learnable prompt \( \pp_c \) that conditions the generator \( G \) to synthesize samples of class \( c^d \). Our objective is to identify a transformation trajectory in the LDM latent space such that, given an arbitrary real image \( \xx \) as input to \( G \), the learned prompt \( \pp_c \) guides \( G \) to generate a dreamed version \( \xx^d \) that shares some characteristics with class \( c \), while remaining distinct enough to not be classified as \( c \) by the model \( F_{\btheta} \). This ensures that the dreamed samples populate a structured latent space region that remains visually coherent but semantically separated from real classes. \\
Formally, we structure the dream class condition $\pp_c = [\pp_{\text{soft}, c}, \pp_{\text{text}, c}]$ with \( \pp_{\text{text}, c} \) being the fixed text prompt describing the transformation for class \( c \), as: ``\texttt{An image of class~} [${c}$]'',
and \( \pp_{\text{soft}, c} \) being a learnable soft prompt vector optimized to refine the conditioning for class \( c \).

\begin{figure}
    \centering
    \includegraphics[width=0.6\linewidth]{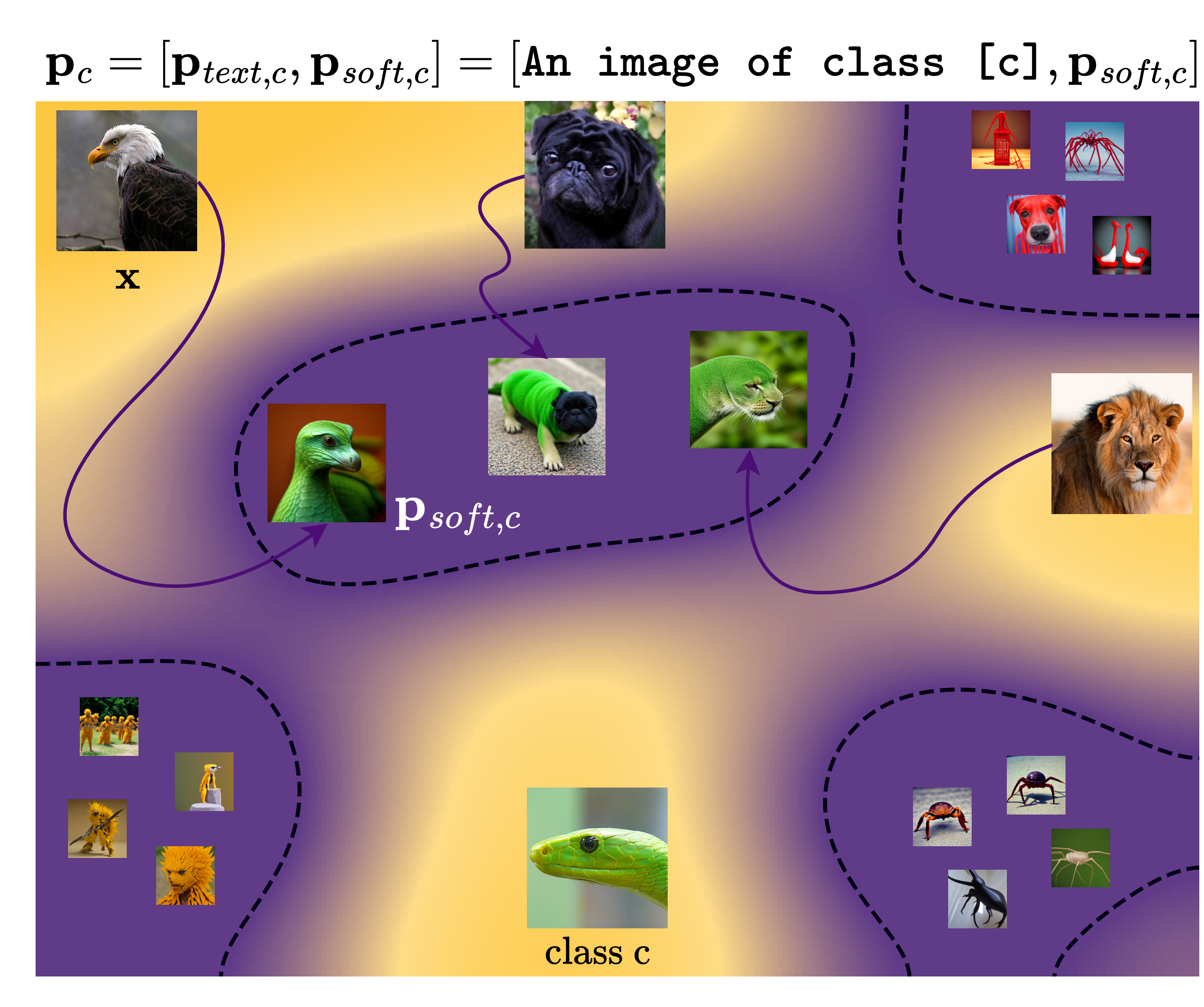}
    \caption{\textbf{Visualization of the dreaming optimization process in the latent space of a LDM.} Given a real sample \( \xx \), the optimization refines the soft prompt \( \pp_{\text{soft},c} \) to steer the diffusion model towards generating a dreamed counterpart that aligns with the target class $c$ (e.g., a green mamba in this example). 
    The dreaming process explores latent regions where images are visually similar yet distinct from target classes, forming novel intermediate classes (violet zones).
    }\label{fig:dreaming_process}
\end{figure}

Due to the stochasticity of \( G \), multiple dreamed samples can be generated from the same real image \( \xx \) and condition \( \pp_c \).  Fig.~\ref{fig:dreaming_process} shows the dreaming optimization process in the LDM latent space.

Prompt optimization is performed by minimizing the cross-entropy loss:
\begin{equation}
\min_{\pp_{\text{soft},c}} \mathbb{E}_{\xx \sim \mathcal{D}_i \setminus \mathcal{D}_i^c} \bigl[ - \log p(c | G (\xx, \pp_c ); \btheta ) \bigr] ,
\label{eq:prompt_optim}
\end{equation}
where \( \mathcal{D}_i^c \) is the subset of real samples belonging to class \( c \), excluded from the optimization process: this ensures that optimization is not conditioned on images already belonging to the target distribution, allowing the process to gradually converge toward it while generating novel yet structured concepts.
As optimization progresses, dreamed samples populate a distinct but structured latent region, allowing future tasks to benefit from enhanced feature reuse and transferability.

\begin{figure*}[ht]
    \centering
    \renewcommand{\thesubfigure}{}
    \setlength{\tabcolsep}{0pt}
    \renewcommand{\arraystretch}{0}

    \begin{tabular}{cccccccccc}
        
        \subfloat{\includegraphics[width=0.095\textwidth]{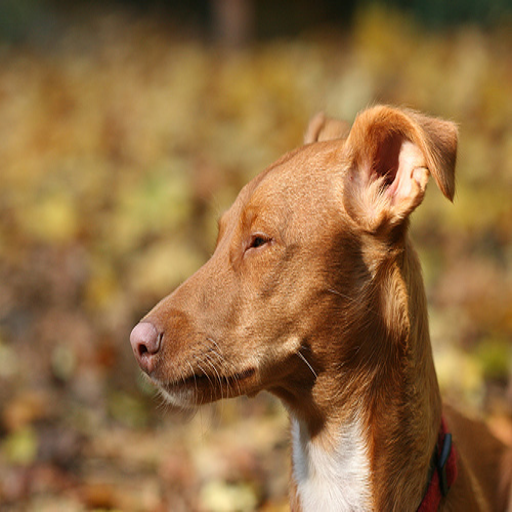}} &
        \subfloat{\includegraphics[width=0.095\textwidth]{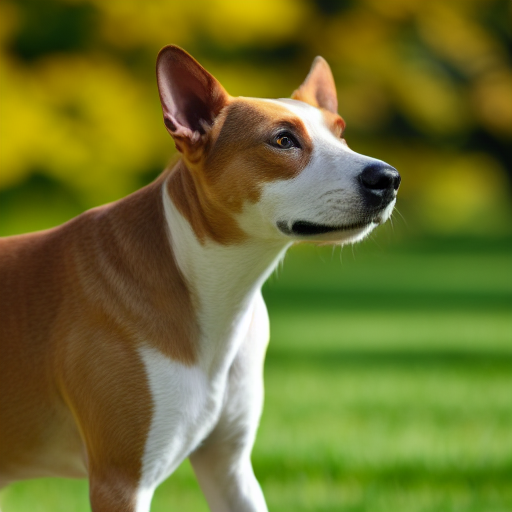}} &
        \subfloat{\includegraphics[width=0.095\textwidth]{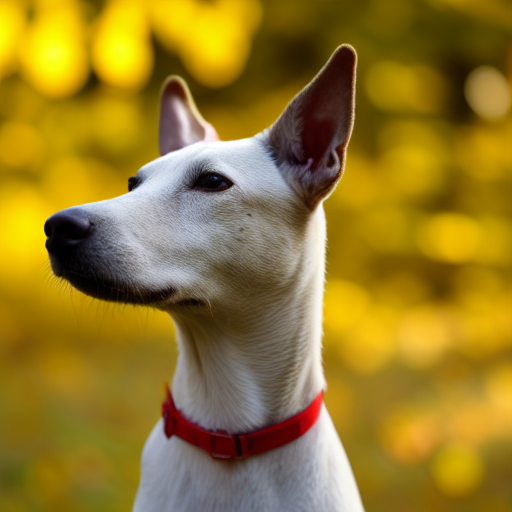}} &
        \subfloat{\includegraphics[width=0.095\textwidth]{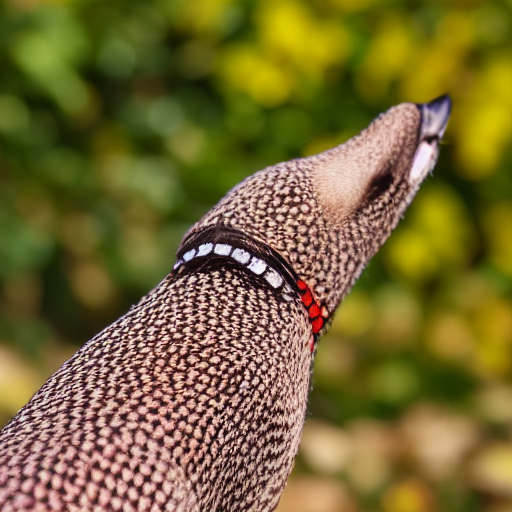}} &
        \subfloat{\includegraphics[width=0.095\textwidth]{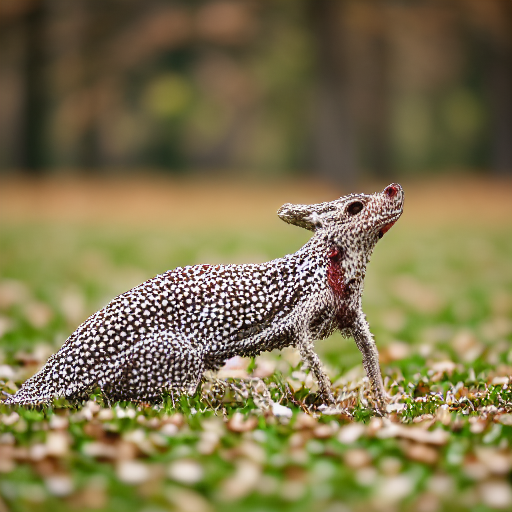}} &
        \subfloat{\includegraphics[width=0.095\textwidth]{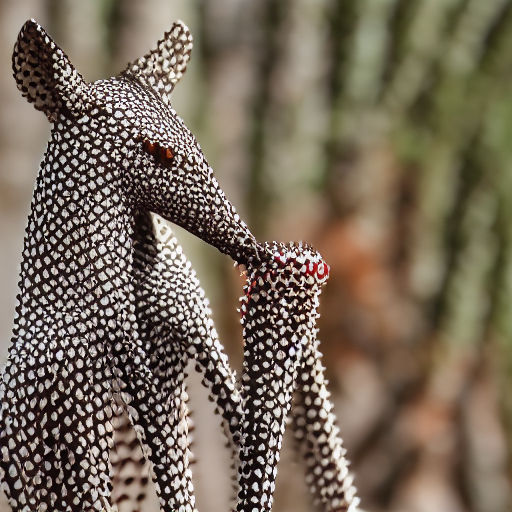}} &
        \subfloat{\includegraphics[width=0.095\textwidth]{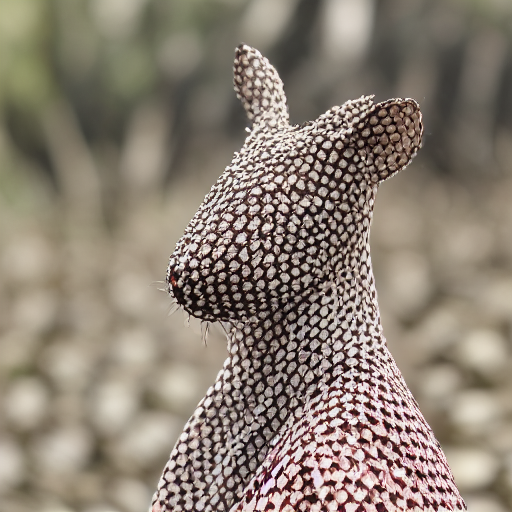}} &
        \subfloat{\includegraphics[width=0.095\textwidth]{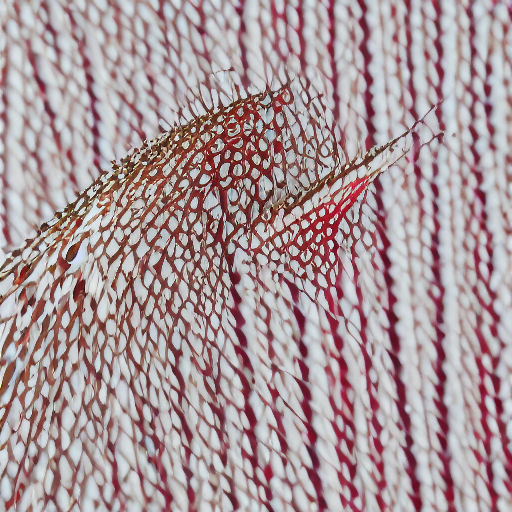}} &
        \subfloat{\includegraphics[width=0.095\textwidth]{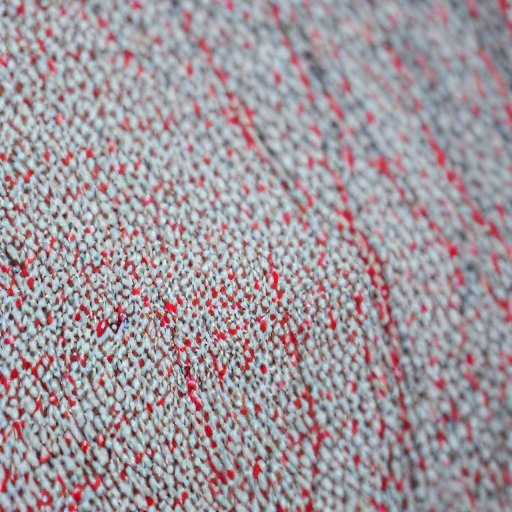}} &
        \subfloat{\includegraphics[width=0.095\textwidth]{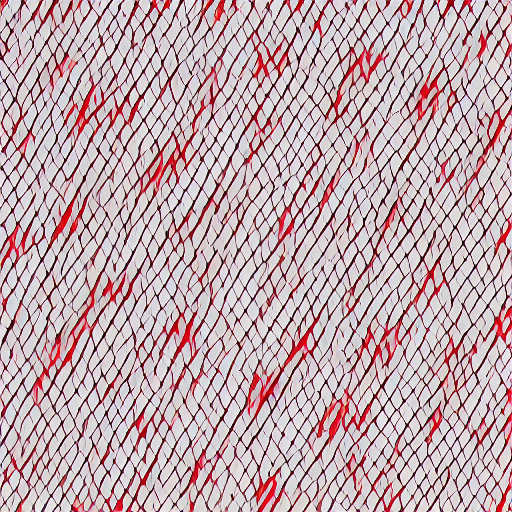}} \\
        
        \subfloat{\includegraphics[width=0.095\textwidth]{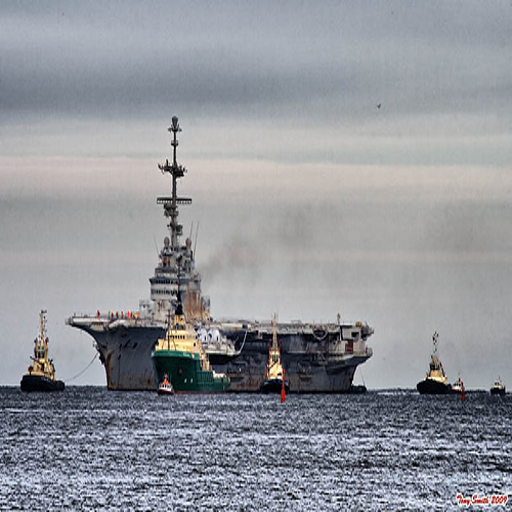}} &
        \subfloat{\includegraphics[width=0.095\textwidth]{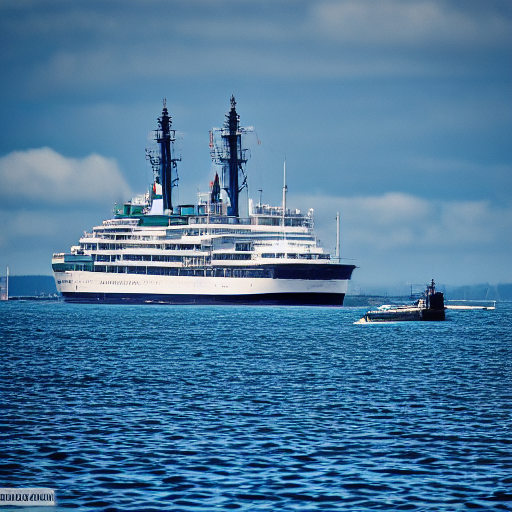}} &
        \subfloat{\includegraphics[width=0.095\textwidth]{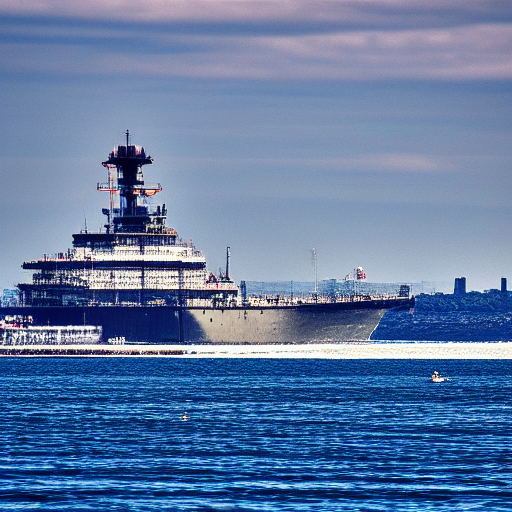}} &
        \subfloat{\includegraphics[width=0.095\textwidth]{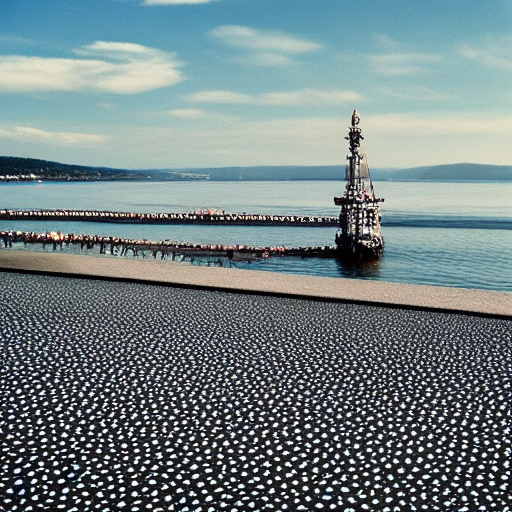}} &
        \subfloat{\includegraphics[width=0.095\textwidth]{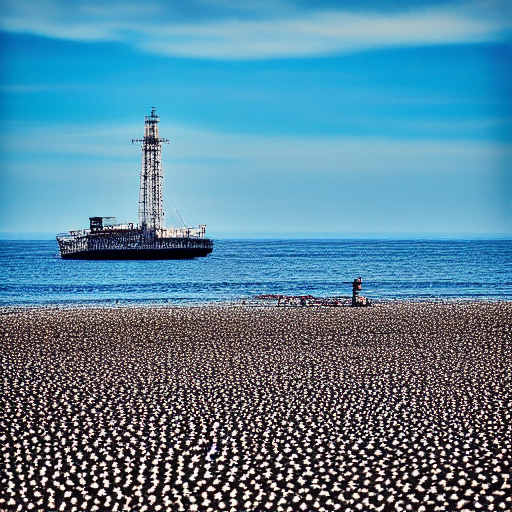}} &
        \subfloat{\includegraphics[width=0.095\textwidth]{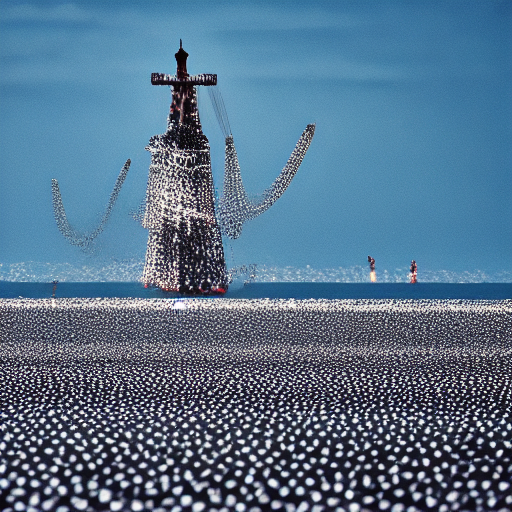}} &
        \subfloat{\includegraphics[width=0.095\textwidth]{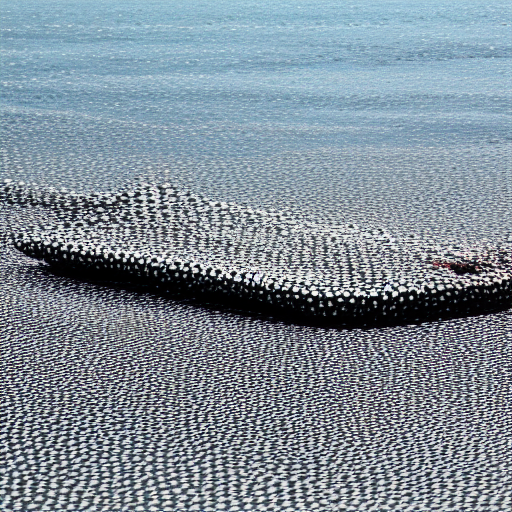}} &
        \subfloat{\includegraphics[width=0.095\textwidth]{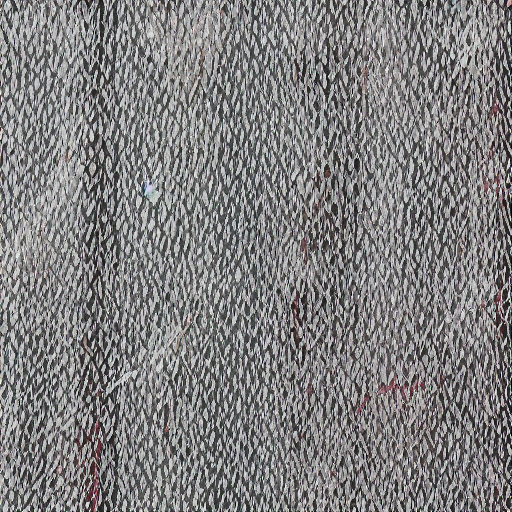}} &
        \subfloat{\includegraphics[width=0.095\textwidth]{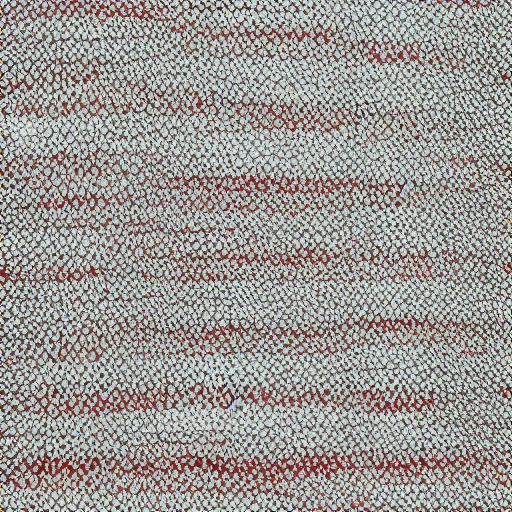}} &
        \subfloat{\includegraphics[width=0.095\textwidth]{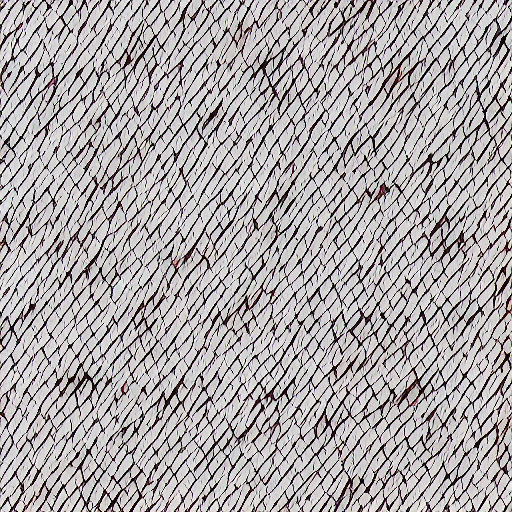}} \\

        \subfloat{\includegraphics[width=0.095\textwidth]{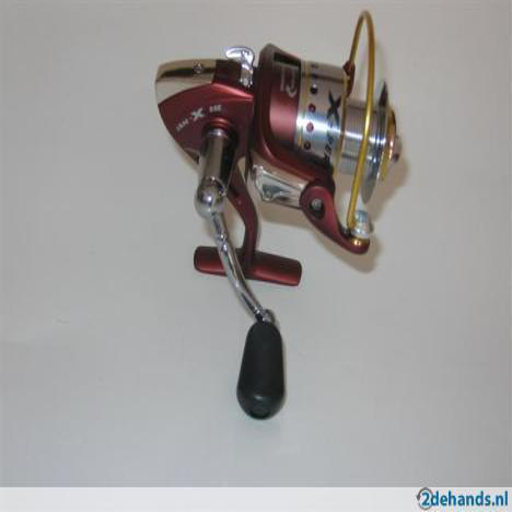}} &
        \subfloat{\includegraphics[width=0.095\textwidth]{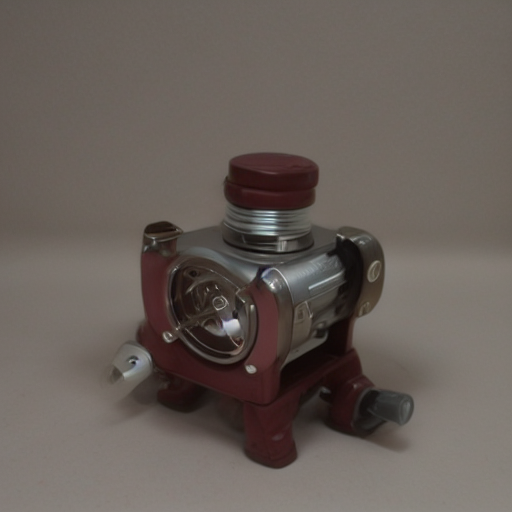}} &
        \subfloat{\includegraphics[width=0.095\textwidth]{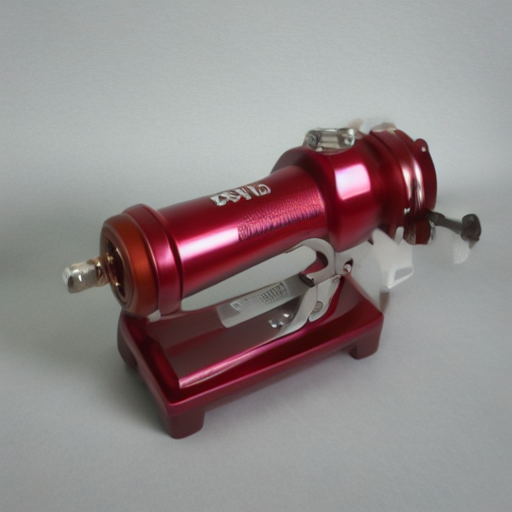}} &
        \subfloat{\includegraphics[width=0.095\textwidth]{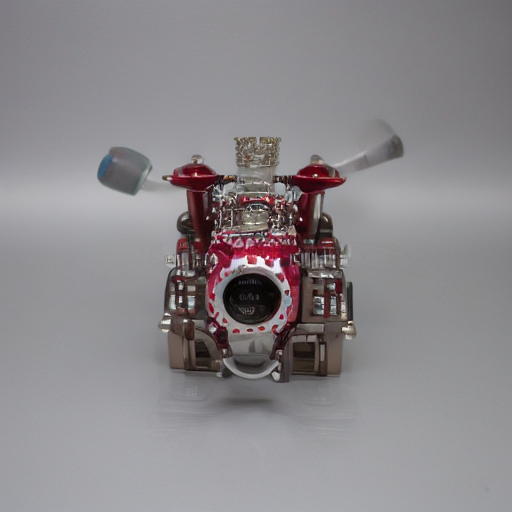}} &
        \subfloat{\includegraphics[width=0.095\textwidth]{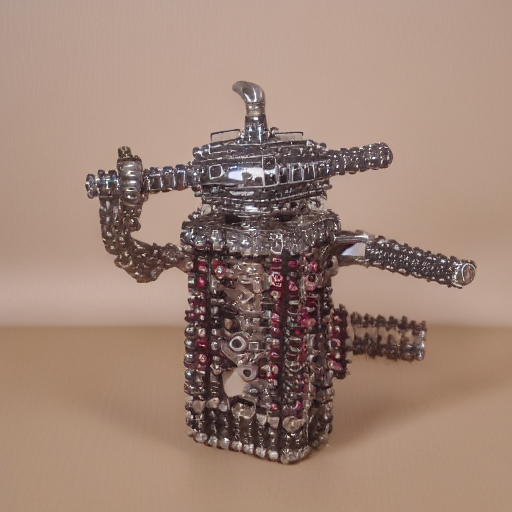}} &
        \subfloat{\includegraphics[width=0.095\textwidth]{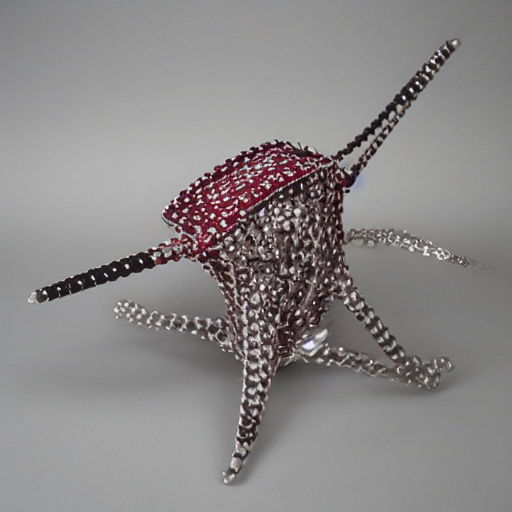}} &
        \subfloat{\includegraphics[width=0.095\textwidth]{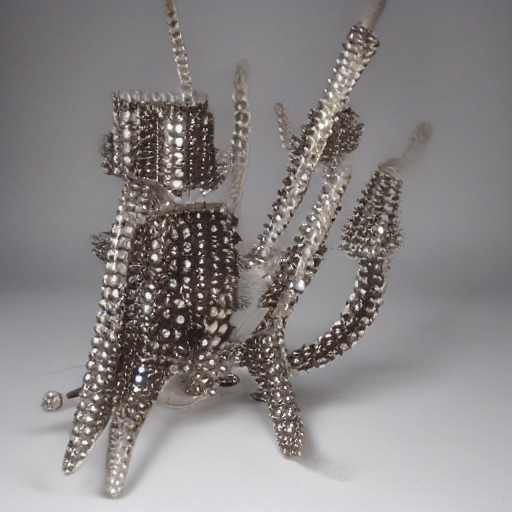}} &
        \subfloat{\includegraphics[width=0.095\textwidth]{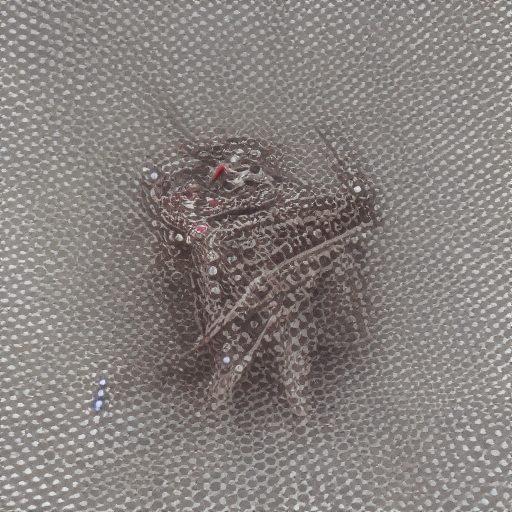}} &
        \subfloat{\includegraphics[width=0.095\textwidth]{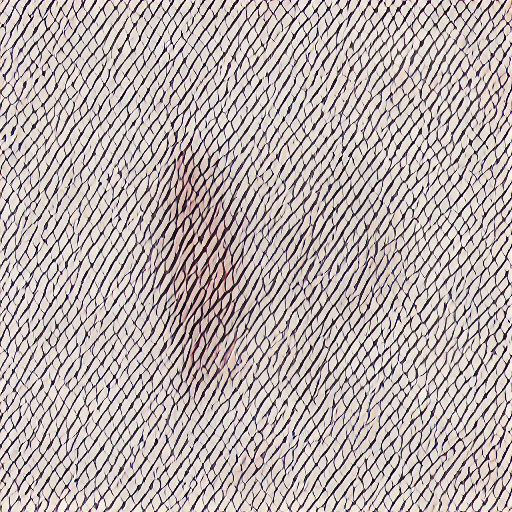}} &
        \subfloat{\includegraphics[width=0.095\textwidth]{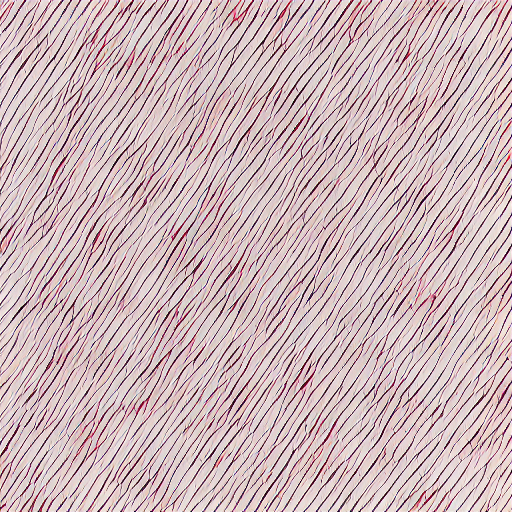}} \\
        
    \end{tabular}
    \caption{\textbf{Examples of dreaming optimization trajectories showing collapse}. From left to right, the images depict different stages of the optimization process. Each row illustrates the evolution of three example images throughout the same prompt optimization. Initially, the generated samples maintain meaningful variations. However, as optimization progresses, they become increasingly similar, reducing diversity and leading to less effective representations.}
    \label{fig:collapse}
    
\end{figure*}

Dreaming optimization produces a set of conditioning prompts $\left\{ \pp_c ~|~ c \in \cC_t \right\}$ for each class of task $\tau_t$. Next, we determine the output neurons to which the new dream classes \edit{$\dD_{t, \cC_t}^{d} = \bigcup_{c} \dD_{t,c}^{d}$} should be mapped, replacing a subset of dream classes from past tasks. The rationale for this step is to map new classes over ``similar'' past dream classes, to ensure a smooth integration and prevent high gradients during training. Thus, let $\cC$ be the set of all output model neurons, and $\cC_{\text{real}} = \bigcup_{i \le t} \cC_i$ the set of outputs assigned to real past tasks. We compute the set of possible destination neurons for the new dream classes as $\cC_{\text{avail}} = \cC \setminus \cC_{\text{real}}$. Then, we obtain the output neuron $c_{\text{out}}$ for dream class $c^d$ as:

\begin{equation}
c_{\text{out}} = \argmin_{c \in \cC_{\text{avail}}} \Ee_{\xx \sim G_{c^d}}  \bigl[ -\log p( c | \xx ; \btheta) \bigr],
\label{eq:c_out_mapping}
\end{equation}
where $G_c$ is the distribution associated to samples produced by $G$ when conditioned with $\pp_c$. In practice, $c^d$ replaces the dream class to which it is more ``aligned'' in terms of classification likelihood.

\label{sec:algorithm}

D2L pipeline is described in Alg.~\ref{alg:d2l}. For clarity of presentation, the process is simplified by omitting the initial and final tasks. The initial task lacks dreaming classes, reducing the training loss to Eq.~\ref{eq:ce_t1}, while the final task does not perform dreaming generation and optimization as these phases are unnecessary.

\algrenewcommand\algorithmicrequire{\textbf{Notation}}
\algrenewcommand\algorithmicensure{\textbf{}}
\begin{algorithm*}[ht!]
\caption{Dream2Learn (D2L)}
    \begin{algorithmic}[1]
        \Require
        \Statex $T$, number of tasks; \quad $\cC_t$, classes of task $t$
        \Statex $F_{\btheta}$, the continual classifier
        \Statex $G$, generator; \quad $\bB$, buffer
        \Statex $\dD_{t}$, the real data distribution at task $t$
        \Statex $\dD_{t-1}^{d}$ the \textit{dream} distribution used during continual training at task $t$
        \Statex $\dD_{t}^r$, the distribution of the dream classes to be \textit{removed} at task $t$
        \Statex $\dD_{t-1}^{d*}$, the residual \textit{dream} classes after mapping at task $t$
        \Statex $\xx$, a real image 
        \Statex $\xx^d$, a generated \textit{dream} image
        \Statex $\pp_c$, the learnable prompt associated with class $c$
        \Statex $\dD_{t, c}^{d}$, the distribution of \textit{dreams} generated after task $t$ from class $c$
        \Statex $\dD_{t, \cC_t}^{d}$, the distribution of \textit{dreams} generated after task $t$ from classes $\cC_t$
        \Statex $\dD_{t}^d$, the distribution of all \textit{dreams} after task $t$
        \Statex \hrulefill
        \For{$t = 2$ \textbf{to} $T-1$}
            \State $\dD_{t}^r \leftarrow Mapping(F_{\btheta}, \dD_{t})$ \Comment{\textit{real classes mapping} \edit{(Sec.~\ref{sec:method})}}
            \State $\dD_{t-1}^{d*} \leftarrow \dD_{t-1}^{d} \setminus \dD_{t}^r$
		      \ForAll {epochs} \Comment{\textit{CL training}}
                \State $loss \leftarrow \mathcal{L}_{CE}(F_{\btheta}, \dD_{t-1}^{d*} \cup \dD_t) +  \lL_{\text{CL}}(F_{\btheta}, \dD_t, \bB)$ \Comment{\edit{Eq.~\ref{eq:ce_tn}}}
                \State $\bB \leftarrow ReservoirSample(\bB, \dD_t)$
                \State \textbf{update} $\btheta$
            \EndFor
            \ForAll{$c \in \cC_t$} \Comment{\textit{dreaming optimization} \edit{(Sec.~\ref{sec:dream})}}
                \Repeat
                    \State $\xx^d \leftarrow G(\xx, \pp_c)$
                    \State $loss \leftarrow \mathcal{L}_{CE}(F_{\btheta}, (\xx^d, c))$ \Comment{\edit{Eq.~\ref{eq:prompt_optim}}}
                    \State \textbf{update} $\pp_c$
                    \State $stop \leftarrow Oracle(\xx, \xx^d)$ \Comment{\edit{Sec.~\ref{sec:oracle}}}
                \Until $stop$
                \State $\dD_{t, c}^{d} \leftarrow Generate(G, \bB, \pp_c)$ \Comment{\textit{dreaming generation} \edit{(Sec.~\ref{sec:dream})}}
            \EndFor
            \State $\dD_{t, \cC_t}^{d} \leftarrow \bigcup_{c} \dD_{t,c}^{d}$
            \State $\dD_{t}^r \leftarrow Mapping(F_{\btheta}, \dD_{t, \cC_t}^{d})$ \Comment{\textit{dream classes mapping} \edit{(Eq.~\ref{eq:c_out_mapping})}}
            \State $\dD_{t}^d \leftarrow (\dD_{t-1}^{d*} \setminus \dD_{t}^r) \cup \dD_{t, \cC_t}^{d}$
		\EndFor
	\end{algorithmic} 
    \label{alg:d2l}
\end{algorithm*}

\subsection{Oracle-guided optimization}
\label{sec:oracle}

One key challenge in the dreaming process is determining when to stop optimizing the soft prompt \( \pp_{\text{soft},c} \) to avoid collapse or excessive task-specific bias, as illustrated in Fig.~\ref{fig:collapse}. If optimization continues indefinitely toward the convergence of Eq.~\ref{eq:prompt_optim}, the generated samples risk becoming redundant or overfitting to the current task, reducing their effectiveness for future learning.
To prevent this, we introduce an oracle network that predicts the optimal stopping point by evaluating whether further refinement of \( \pp_{\text{soft},c} \) contributes to meaningful latent representation learning. The oracle is trained on a separate dataset $\dD_O$, where stopping decisions are labeled based on the quality of generated dreams. \\
Formally, we define the oracle network  \( O \), which takes as input a sequence of feature vectors extracted over a temporal window of  \( k \)  optimization steps and provides a binary decision:
\begin{equation}
O(\mathbf{Z}_t) \in \left\{0,1\right\}
\end{equation}
where \( \mathbf{Z}_t \) is the aggregated feature matrix over the last \( k \) generated samples:
\begin{equation}
\mathbf{Z}_t = \left[ \zz_{t-k+1}, \zz_{t-k+2}, \dots, \zz_t \right]
\end{equation}
The components of each vector \( \zz_i \in \Re^4 \) are the following quantities, computed using the generator $G_c$ conditioned by $\pp_{\text{soft},c}$ at the $i$-th optimization iteration of Eq.~\ref{eq:prompt_optim}: 
1) \( \Ee_{\xx } \bigl[ \text{sim}(\xx, G_c(\xx) \bigr] \), with $\text{sim}(\cdot)$ measuring the structural similarity between the generated image \( G_c(\xx) \) and its conditioning image \( \xx \), ensuring that the generated image maintains structural coherence; 
2) \( \Ee_{\xx } \bigl[ f_{\btheta}(\xx) ^\intercal f_{\btheta}\bigl(G_c(\xx)\bigr) \bigr] \), i.e., the dot product between feature embeddings $f$ extracted by the classifier \( F_{\btheta} \), capturing the alignment between the representations of \( \xx \) and \( G_c(\xx) \); 
3) \( \Ee_{\xx } \bigl[ \text{Q}\bigl(G_c(\xx)\bigr) \bigr] \), where $Q$ computes the CLIP-based Image Quality Assessment~\citep{wang2023exploring}, evaluating the perceptual quality of the generated image;
4) \( \Ee_{\xx } \Bigl[ \sigma\Bigl( f_{\btheta}\bigl(G_c(\xx)\bigr) \Bigr) \Bigr] \), i.e., the standard deviation of the feature embeddings, capturing the diversity within generated samples. 
The optimization process halts once the oracle outputs 1 for \( n \) consecutive iterations, ensuring robustness to fluctuations in individual predictions. 
Once trained, the oracle network \( O \) is frozen and used across all tasks to determine when to stop the optimization of \( \pp_{\text{soft},c} \), ensuring that the dreaming process remains effective without collapsing. 
\section{Experimental Results}

\subsection{Benchmarks}
\label{sec:benchmarks}
We evaluate D2L on three continual learning benchmarks, obtained by splitting image classification datasets into a series of disjoint tasks:

\begin{itemize}[noitemsep,nolistsep,leftmargin=*]
    \item \textbf{Split Mini-ImageNet}~\citep{vinyals2016matching}: a widely used few-shot learning dataset, consisting of ImageNet 100 classes with 600 samples each, commonly adapted for continual learning;
    \item \textbf{Split FG-ImageNet}~\citep{russakovsky2015imagenet}\footnote{Split\ FG-ImageNet is derived from \url{https://www.kaggle.com/datasets/ambityga/imagenet100}}, 
    a fine-grained image classification benchmark with 100 animal classes from ImageNet, designed to evaluate continual learning methods on a more challenging task.
    \item \textbf{Split ImageNet-R}~\citep{hendrycks2021many} comprises various renditions of 200 ImageNet classes (e.g., paintings, sculptures, embroidery, cartoons, etc.), with an average of 150 samples per class and a class-imbalanced distribution, introducing strong intra-class variations. 
\end{itemize}

In our experimental setup, half of the classes in each dataset are used for the first task, while the remaining classes are equally split across the subsequent tasks. In particular, excluding the first task, the Mini-ImageNet and FG-ImageNet datasets consist of 10 tasks with 5 classes each, whereas the ImageNet-R dataset consists of 5 tasks with 20 classes each.\\
\subsection{Evaluation and training procedure}\label{sec:training}
\noindent \textbf{Evaluation protocol.} All experiments are conducted under the standard 
\emph{training-from-scratch CNN} regime commonly adopted in rehearsal-based class-incremental learning. The backbone is randomly initialized and trained sequentially without any external pretraining. We use ResNet-18~\citep{he2016deep}, which has become a widely used benchmark architecture for controlled comparison of continual learning strategies under fixed model capacity and optimization settings. A distinct line of continual learning research instead builds upon large pre-trained models, either via full or partial fine-tuning~\citep{ramasesh2021effect, boschini2022transfer, pmlr-v199-ostapenko22a} or through prompt-tuning approaches~\citep{Wang_2022_CVPR, smith2023coda} 
that adapt frozen Vision Transformer (ViT) backbones. These methods operate under a different experimental regime, where learning starts from representations shaped by large-scale pretraining.
In the present study, the classifier is trained entirely from scratch, and its learned representations directly guide the dream generation process. Accordingly, all comparisons are performed within this same training-from-scratch CNN setting.\\

\noindent \textbf{Training procedure.} The ResNet-18~\citep{he2016deep} is trained for 10 epochs per task using SGD (learning rate 0.03, batch size 32). Given the large number of classes, we use buffer sizes of 2000 and 5000.\\
Prompt optimization is performed in Stable Diffusion’s text space via cross-entropy loss, guided by classifier predictions. We use Adam (learning rate 0.1, batch size 1), with the number of iterations controlled by the oracle network.  In particular, the oracle is trained once on labeled dreaming trajectories generated from a dataset composed of 100 ImageNet classes disjoint from all evaluation benchmarks (Mini-ImageNet, FG-ImageNet, and ImageNet-R). 
Each trajectory is obtained by optimizing the soft prompt $\mathbf{p}_{\text{soft},c}$ for up to 500 iterations. The stopping label corresponds to the earliest iteration at which generated samples exhibit high perceptual quality and sufficient diversity, while avoiding collapse or overspecialization. At iteration $t$, the oracle receives a temporal feature sequence $\mathbf{Z}_t = [\mathbf{z}_{t-k+1}, \dots, \mathbf{z}_t]$, where each $\mathbf{z}_i \in \mathbb{R}^4$ summarizes generation dynamics. 
From an initial pool of 25 candidate metrics (defined in Appendix~\ref{sec:oracle_features}), spanning image-level quality measures, feature-level statistics, and classifier-based uncertainty signals, four features are selected via SHAP-based importance analysis. The oracle is implemented as a lightweight MLP with one hidden layer (32 units, ReLU) and sigmoid output. Training samples are constructed by sliding a window of length $k=3$ over each trajectory. The model is trained with Adam (learning rate 0.001) using binary cross-entropy loss for up to 500 epochs, with early stopping on a validation split. 
Once trained, the oracle is frozen and reused across all tasks and datasets. During prompt optimization, termination is triggered when a stop signal is predicted in at least $n=2$ of the past $k=3$ iterations.
All experiments are implemented on top of the Mammoth framework~\cite{buzzega2020rethinking,buzzega2020dark,bellitto2022effects,caccia2022new,fini2022self,sorrenti2023,sorrenti2023selective}, and results are reported in the class-incremental setting as mean $\pm$ 
standard deviation over 5 runs.

\subsection{Results}
\label{sec:results}
Since the D2L mechanism leverages prompts optimized 
from representations learned through rehearsal and relies on  past samples stored in a buffer, we integrate it specifically with rehearsal-based class-incremental methods. In this setting, D2L functions as an orthogonal augmentation to rehearsal, enabling 
us to isolate its contribution without introducing architectural or optimization confounds. We therefore evaluate it in conjunction with DER++~\citep{buzzega2020dark}, ER-ACE~\citep{caccia2022new}, 
and ER~\citep{chaudhry2019tiny}, comparing their performance with 
and without the proposed dreaming mechanism.

Tab.~\ref{tab:results} presents results in terms of \emph{final average accuracy (FAA)} in the class-incremental setting, i.e., the accuracy on a separate test set including all task classes, measured after training on the last task, with no knowledge on task identity at inference time. Our approach leads to a significant improvement in performance across all benchmarks, demonstrating the importance of mimicking human dreaming for mitigating forgetting.
\begin{table*}[htb]
    \centering
    \renewcommand{\arraystretch}{1.2}
        \caption{\textbf{Class-incremental final average accuracy (\textbf{FAA}) of rehearsal-based methods}, with and without dreaming, for buffer sizes 2000 and 5000.}
    \footnotesize {
    \begin{tabular}{l|cc|cc|cc}
        \toprule
         & \multicolumn{2}{c}{{\textbf{Mini-ImageNet}}} & \multicolumn{2}{c}{{\textbf{FG-ImageNet}}}      & \multicolumn{2}{c}{{\textbf{ImageNet-R}}}\\
        \midrule
          & \multicolumn{1}{c}{\textbf{2000}}  & \multicolumn{1}{c}{\textbf{5000}}  & \multicolumn{1}{c}{\textbf{2000}}  & \multicolumn{1}{c}{\textbf{5000}} &\multicolumn{1}{c}{\textbf{2000}}  & \multicolumn{1}{c}{\textbf{5000}}\\
        \cmidrule(lr){1-1} \cmidrule(lr){2-3} \cmidrule(lr){4-5} \cmidrule(lr){6-7}
        ER                             & \resultnof{27.91}{3.49}   & \resultnof{34.21}{3.04}                & \resultnof{21.08}{2.38}   &\resultnof{22.21}{3.44}                & \resultnof{7.68}{0.97}   & \resultnof{10.69}{1.29}\\
        \rowcolor{gray!10}
        \hspace{0.1 cm} {$\hookrightarrow$\textbf{D2L}}    & \resultnof{\textbf{31.18}}{2.74}   & \resultnof{\textbf{39.75}}{2.61}                & \resultnof{\textbf{23.53}}{1.98}      & \resultnof{\textbf{32.73}}{3.39}                   & \resultnof{\textbf{8.67}}{0.66}    & \resultnof{\textbf{11.84}}{0.95}\\
        DER++                            & \resultnof{14.74}{2.14}   & \resultnof{26.92}{4.72}                & \resultnof{14.43}{3.68}   & \resultnof{23.86}{2.54}                & \resultnof{6.08}{0.81}   & \resultnof{8.29}{1.15}\\
        \rowcolor{gray!10}
        \hspace{0.1 cm} {$\hookrightarrow$\textbf{D2L}}    & \resultnof{\textbf{21.06}}{5.45}   & \resultnof{\textbf{31.91}}{5.19}                & \resultnof{\textbf{18.86}}{3.22}      & \resultnof{\textbf{25.38}}{2.17}                   & \resultnof{\textbf{8.60}}{1.00}    & \resultnof{\textbf{10.89}}{1.56}\\
        \arrayrulecolor{gray}
        \arrayrulecolor{black}
        ER-ACE                & \resultnof{33.26}{3.51}    & \resultnof{40.59}{1.20}                & \resultnof{24.79}{5.02}   & \resultnof{30.16}{4.97}                & \resultnof{7.09}{0.59}   & \resultnof{9.44}{0.70}\\
        \rowcolor{gray!10}
        \hspace{0.1 cm} {$\hookrightarrow$\textbf{D2L}}    & \resultnof{\textbf{40.90}}{0.95}   & \resultnof{\textbf{47.32}}{0.89}                & \resultnof{\textbf{31.57}}{1.20}      & \resultnof{\textbf{38.50}}{1.01}                   & \resultnof{\textbf{9.54}}{0.39}    & \resultnof{\textbf{12.51}}{0.56}\\
        \bottomrule
    \end{tabular}}
    \label{tab:results}
\end{table*}

\begin{table*}[htb!]
    \centering
    \renewcommand{\arraystretch}{1.2}
        \caption{\textbf{Forward Transfer (FWT) of rehearsal-based methods}, with and without dreaming, for buffer sizes 2000 and 5000.}

    \footnotesize {
    \begin{tabular}{l|cc|cc|cc}
        \toprule
                                    & \multicolumn{2}{c}{{\textbf{Mini-ImageNet}}} & \multicolumn{2}{c}{{\textbf{FG-ImageNet}}}      & \multicolumn{2}{c}{{\textbf{ImageNet-R}}}\\
        \midrule
           & \multicolumn{1}{c}{\textbf{2000}}  & \multicolumn{1}{c}{\textbf{5000}}  & \multicolumn{1}{c}{\textbf{2000}}  & \multicolumn{1}{c}{\textbf{5000}} &\multicolumn{1}{c}{\textbf{2000}}  & \multicolumn{1}{c}{\textbf{5000}}\\
        
        \cmidrule(lr){1-1} \cmidrule(lr){2-3} \cmidrule(lr){4-5} \cmidrule(lr){6-7}
        
        ER                           & -2.58     & -1.62             & -1.88     & -1.52             & -1.32     & -0.64\\
        \rowcolor{gray!10}
        \hspace{0.1 cm} {$\hookrightarrow$\textbf{D2L}}    & \textbf{+0.33} & \textbf{+0.47} & \textbf{+1.79}   & \textbf{+1.19}             & \textbf{+0.24}        & \textbf{+0.14}\\
        
        DER++                       & -1.55     & -2.00             & -1.36     & -2.48             & -1.03     & -1.83\\
        \rowcolor{gray!10}
        \hspace{0.1 cm} {$\hookrightarrow$\textbf{D2L}}    & \textbf{+0.97}    & \textbf{+0.71}             & \textbf{-0.13}    & \textbf{+1.86}             & \textbf{+0.30}    & \textbf{+0.24}\\
        \arrayrulecolor{gray}
                                                                                                        
        \arrayrulecolor{black}
        
        ER-ACE                         & -1.99     & -2.45             & -2.00     & -2.16             & -2.46     & -1.27\\
        \rowcolor{gray!10}
        \hspace{0.1 cm} {$\hookrightarrow$\textbf{D2L}}    & \textbf{+1.05}   & \textbf{-1.58}        & \textbf{+1.09}     & \textbf{+0.08}        & \textbf{-0.25}   & \textbf{+0.17}\\
        
        \bottomrule
    \end{tabular}}
    \label{tab:fwt}
\end{table*}

\begin{table*}[h!]
    \centering
    \caption{\textbf{Comparison with state-of-the-art methods, in terms of class-incremental final average accuracy (FAA)}, for  buffer sizes 2000 and 5000.}
    \label{tab:competitors}
    \rowcolors{15}{white}{gray!10}
    \renewcommand{\arraystretch}{1.2}
    \footnotesize {
    
    \begin{tabular}{l|cc|cc|cc}
        \toprule
        \textbf{Method}                                     &\multicolumn{2}{c}{\textbf{Mini-ImageNet}}                         &\multicolumn{2}{c}{\textbf{FG-ImageNet}}                      & \multicolumn{2}{c}{\textbf{ImageNet-R}}\\
        \midrule
        \rowcolor{gray!10}
        Fine-tune                                           & \multicolumn{2}{c}{\resultnof{6.72}{1.20}}                           & \multicolumn{2}{c}{\resultnof{6.98}{0.10}}                       & \multicolumn{2}{c}{\resultnof{4.46}{0.15}}\\ 
        \midrule
        \multicolumn{1}{c}{\textbf{}} & \multicolumn{6}{c}{\textbf{Buffer-based methods}}\\
        \midrule
                                     & \multicolumn{1}{c}{\textbf{2000}}    & \multicolumn{1}{c}{\textbf{5000}}        & \multicolumn{1}{c}{\textbf{2000}}  & \multicolumn{1}{c}{\textbf{5000}}       & \multicolumn{1}{c}{\textbf{2000}}  & \multicolumn{1}{c}{\textbf{5000}}\\
        \cmidrule(lr){1-1} \cmidrule(lr){2-3} \cmidrule(lr){4-5} \cmidrule(lr){6-7}
        GSS                   & \resultnof{6.40}{0.38}     & \resultnof{5.71}{0.08}                 & \resultnof{8.07}{0.26}   & \resultnof{9.23}{0.85}        & \resultnof{5.08}{0.13}   & \resultnof{4.29}{0.32}\\
        \rowcolor{gray!10}
        A-GEM               & \resultnof{6.78}{1.13}     & \resultnof{7.45}{0.76}         & \resultnof{6.20}{1.11}   & \resultnof{6.11}{1.13}        & \resultnof{4.69}{0.03}   & \resultnof{6.29}{0.84}\\
        RPC                      & \resultnof{9.22}{0.30}     & \resultnof{9.02}{0.24}         & \resultnof{8.13}{0.11}   & \resultnof{7.41}{0.74}        & \resultnof{5.71}{0.03}   & \resultnof{6.32}{0.80}\\
        \rowcolor{gray!10}
        DER++                     & \resultnof{14.74}{2.14}    & \resultnof{26.92}{4.72}        & \resultnof{14.43}{3.68}  & \resultnof{23.86}{2.54}       & \resultnof{6.08}{0.81}   & \resultnof{8.29}{1.15}\\
        FDR                 & \resultnof{15.46}{1.10}    & \resultnof{11.58}{0.96}        & \resultnof{9.17}{2.40}   & \resultnof{12.91}{0.95}       & \resultnof{5.71}{0.18}   & \resultnof{5.77}{0.10}\\
        \rowcolor{gray!10}
        iCaRL                    & \resultnof{16.46}{0.51}    & \resultnof{16.50}{0.33}        & \resultnof{8.54}{0.88}   & \resultnof{8.86}{0.25}        & \resultnof{1.97}{0.28}   & \resultnof{1.91}{0.29}\\
        ER                          & \resultnof{27.91}{3.49}    & \resultnof{34.21}{3.04}        & \resultnof{21.08}{2.38}  & \resultnof{22.21}{3.44}       & \resultnof{7.68}{0.97}   & \resultnof{10.69}{1.29}\\
        BiC                           & \resultnof{30.56}{7.41}    & \resultnof{37.84}{0.61}        & \resultnof{27.83}{2.75}           & \resultnof{32.29}{0.70}                & \resultnof{7.15}{1.14}   & \resultnof{8.60}{2.07}\\
        ER-ACE                 & \resultnof{33.26}{3.51}    & \resultnof{40.59}{1.20}        & \resultnof{24.79}{5.02}  & \resultnof{30.16}{4.97}       & \resultnof{7.09}{0.59}   & \resultnof{9.44}{0.70}\\
        \midrule
        \textbf{ER-ACE + D2L} & \resultnof{\textbf{40.90}}{0.95}     & \resultnof{\textbf{47.32}}{0.89}         & \resultnof{\textbf{31.57}}{1.20}   & \resultnof{\textbf{38.50}}{1.01}        & \resultnof{\textbf{9.54}}{0.39}   & \resultnof{\textbf{12.51}}{0.56}\\
       
        \arrayrulecolor{black}
        \arrayrulecolor{black}
        \bottomrule
    \end{tabular}
    }
\end{table*}

One of our key claims is that our dreaming mechanism enhances a model’s ability to prepare for future tasks. To validate this, we evaluate forward transfer (FWT)~\citep{lopez2017gradient}, measuring how well the model leverages prior knowledge when learning new tasks. FWT is defined as the average difference between the accuracy on a task \( \tau_t \) by a model trained up to \( \tau_{t-1} \), and the accuracy on \( \tau_t \) by a randomly initialized model. Since a continually trained model often predicts known classes, FWT is typically negative. 
Tab.~\ref{tab:fwt} shows that dream generation improves FWT, with D2L achieving positive forward transfer in some cases, similar to WSCL. 
However, WSCL achieves positive forward transfer by relying on additional real data to simulate dreams, whereas D2L internally generates these dreams by leveraging the model’s own internal knowledge.

We further conduct a comprehensive performance analysis by comparing the best performing version of our approach from Tab.~\ref{tab:results} (i.e., ER-ACE + D2L) with state-of-the-art continual learning methods\footnote{Results were obtained using the original code released alongside the corresponding papers.}: 
\edit{GSS~\citep{aljundi2019gradient}, A-GEM~\citep{chaudhry2019efficient},  iCaRL~\citep{rebuffi2017icarl}, FDR~\citep{benjamin2018measuring}, BiC~\citep{wu2019large}, and RPC~\citep{pernici2021class}.}
Results are shown in Tab.~\ref{tab:competitors}. To contextualize these results, we also define a lower bound as simple fine-tuning of the backbone for each task without any countermeasure to forgetting (named \emph{Fine-tune} in the table). 
ER-ACE + D2L outperforms state-of-the-art methods across all examined datasets and buffer sizes, with significant margins.

\subsection{Model analysis}
Model analysis is primarily conducted to assess the contribution of the dream generation strategy. The ER-ACE model~\citep{chaudhry2019tiny}, identified as the top-performing method when combined with our approach (see Tab.~\ref{tab:results}), is used as the baseline model for this study. All experiments are performed on the Mini-ImageNet dataset~\citep{vinyals2016matching}.

\begin{table*}[t]
    \centering
    \caption{\textbf{Ablation on the oracle.} Results on Mini-ImageNet comparing our method with Fixed optimization.}
    \label{tab:ablation_oracle}
    \footnotesize
    
    \begin{tabular}{l|cc}
      \toprule
      & \multicolumn{2}{c}{\textbf{Buffer size}} \\
      \cmidrule(lr){2-3}
      \multirow{-2}{*}{\textbf{Method}} & \textbf{2000} & \textbf{5000} \\
      \midrule
      \rowcolor{gray!10}
      ER-ACE & \resultnof{33.26}{3.51} & \resultnof{40.59}{1.20} \\
      \hspace{4pt}$+$ Fixed optim. & \resultnof{38.94}{0.97} & \resultnof{46.41}{0.55} \\
      \rowcolor{gray!10}
      \hspace{4pt}$+$ \textbf{Oracle  (D2L)} & \resultnof{40.90}{0.95} & \resultnof{47.32}{0.89} \\
      \bottomrule
    \end{tabular}
\end{table*}

\begin{table*}[t]
    \centering
    
    \caption{\edit{\textbf{Impact of dream class updates.} Comparison on Mini-ImageNet of different dreaming strategies.}}
    
    \label{tab:ablation_dreaming_impact}
    \footnotesize
    \begin{tabular}{l|cc}
      \toprule
      & \multicolumn{2}{c}{\textbf{Buffer size}} \\
      \cmidrule(lr){2-3}
      \multirow{-2}{*}{\textbf{Dreaming}} & \textbf{2000} & \textbf{5000} \\
      \midrule
      No dreams & \resultnof{33.36}{3.51} & \resultnof{40.59}{1.20} \\
      \rowcolor{gray!10}
      At beginning & \resultnof{36.93}{2.09} & \resultnof{41.59}{2.54} \\
      Incremental & \resultnof{36.85}{1.16} & \resultnof{43.87}{2.90} \\
      \rowcolor{gray!10}
      \textbf{D2L} & \resultnof{\textbf{40.90}}{0.95} & \resultnof{\textbf{47.32}}{0.89} \\
      \bottomrule
    \end{tabular} 
\end{table*}

\noindent \textbf{Oracle-based optimization.} We first evaluate the contribution of the oracle by comparing the full 
D2L model with a variant without oracle-based stopping, denoted as 
\textit{Fixed optimization}. In this variant, prompt updates are halted 
once the classifier predicts the target class for four consecutive steps, 
using a heuristic termination rule. As shown in Tab.~\ref{tab:ablation_dreaming_type}, incorporating the oracle consistently improves performance. This indicates that adaptive stopping based on generation dynamics enables the dream optimization process to 
better align structural coherence, semantic consistency, and diversity, 
thereby enhancing its contribution to continual learning.\\
We also test the generalization capability of the oracle across 
benchmarks. Specifically, we compare the stopping iterations predicted 
by the single shared oracle, trained as reported in Sect.~\ref{sec:training}, with those obtained using benchmark-specific 
oracles trained separately for Mini-ImageNet, FG-ImageNet, and 
ImageNet-R. The mean absolute deviation between the predicted stopping 
points is 9.28 iterations (out of 500 optimization steps). Importantly, 
no trajectory collapse is observed and continual learning performance 
remains unchanged. These results indicate that the oracle captures 
dataset-agnostic properties of generation dynamics and generalizes 
robustly across benchmarks without requiring retraining.\\
\noindent \textbf{Dream update process.} We next analyze how different strategies for updating dream classes 
during sequential learning affect performance. Our default strategy (D2L) generates a new set of dream classes at the end of each task, matching the number of classes in that task, 
and replaces an equal number of previously generated dreams. 
This keeps the classifier output dimensionality constant 
throughout training. We compare this with three alternative strategies: 1) \emph{No dreams}, where no generated classes are used; 2) \emph{At beginning only}, where dream classes are generated once after the first task and then kept fixed for all subsequent tasks; 3) \emph{Incremental}, where newly generated dream classes are 
accumulated across tasks without replacement, progressively expanding 
the classifier output layer. In this case, new output neurons are 
initialized from $\mathcal{N}(\mu_{\mathbf{w}}, \sigma_{\mathbf{w}}^2)$ 
computed from existing classifier weights. Results in Tab.~\ref{tab:ablation_dreaming_impact} show that the 
proposed replacement strategy consistently achieves the best performance. 
Keeping dreams fixed reduces their relevance to later tasks, while 
uncontrolled expansion increases the dimensionality of the prediction 
space, making optimization more challenging.\\

\noindent  \textbf{Alternative dream generation strategies.} We compare D2L with alternative mechanisms for generating surrogate 
representations, including interpolation-based strategies and 
auxiliary-data approaches. Interpolation-based variants replace dream classes with blended 
representations obtained via: a) \emph{Mixup}~\citep{mixup}, 
combining stream and buffer samples; b) \emph{Continual Mixup}, 
applied only within the replay buffer; c) \emph{Textual Mixup}, 
interpolating class prompt embeddings; and d) \emph{Synth Mixup}, 
interpolating directly in the diffusion latent space.  We also include WSCL~\citep{sorrenti2023}, which leverages an auxiliary 
real dataset to pre-activate future task heads. Results in Tab.~\ref{tab:ablation_dreaming_type} show that D2L 
outperforms all interpolation-based strategies, indicating that 
linear blending of existing representations is insufficient to 
induce the same degree of task-aware restructuring. WSCL achieves 
higher absolute accuracy, reflecting the benefit of additional 
external supervision that provides explicit future-relevant signals. Conceptually, WSCL augments learning through auxiliary real data, 
whereas D2L generates surrogate classes directly from the classifier’s 
internal representations without relying on external datasets. 
Despite this difference in supervision, the two approaches achieve 
comparable performance across buffer sizes. These results suggest 
that internally generated, classifier-guided surrogate classes can 
approximate the benefits of auxiliary-data augmentation while 
operating in a fully self-contained setting.\\

\begin{table*}[ht]
  \centering
  \renewcommand{\arraystretch}{1.2}
  \caption{\edit{\textbf{Comparison of alterative dream generation strategies.} Evaluation on Mini-ImageNet comparing interpolation-based baselines and WSCL (auxiliary-data surrogate against our proposed D2L.}}
  \label{tab:ablation_dreaming_type}
  \footnotesize
  \begin{tabular}{l|cc}
    \toprule
    & \multicolumn{2}{c}{\textbf{Buffer size}} \\
    \cmidrule(lr){2-3}
    \multirow{-2}{*}{\textbf{Method}} & \textbf{2000} & \textbf{5000} \\
    \midrule
    ER-ACE & \resultnof{33.26}{3.51} & \resultnof{40.59}{1.20} \\
    \rowcolor{gray!10}
    \hspace{4pt}$+$ Mixup & \resultnof{36.84}{0.94} & \resultnof{44.82}{1.27} \\
    \hspace{4pt}$+$ Continual Mixup & \resultnof{36.05}{1.22} & \resultnof{43.45}{0.83} \\
    \rowcolor{gray!10}
    \hspace{4pt}$+$ Textual Mixup & \resultnof{31.89}{0.50} & -- \\
    \hspace{4pt}$+$ Synth Mixup & \resultnof{36.99}{0.26} & -- \\
    \rowcolor{gray!10}
    \midrule
    \textbf{D2L} & \resultnof{40.90}{0.95} & \resultnof{47.32}{0.89} \\
    \midrule
    \textbf{WSCL} & \resultnof{\textbf{42.38}}{1.16} & \resultnof{\textbf{48.30}}{2.60} \\
    \bottomrule
  \end{tabular}
\end{table*}

\noindent \textbf{Dream independence from target classes.} Since D2L relies on a pre-trained diffusion model to generate dreamed samples, it is important to verify that the resulting images do not implicitly encode future task classes due to prior exposure during 
diffusion pretraining. We therefore assess whether dreamed samples are statistically out-of-distribution (OOD) with respect to the target benchmark. To this end, we generate dreams on Mini-ImageNet and classify each generated image using a ViT-B/16 model pre-trained on ImageNet-1K, which serves as an external classifier independent of the training 
process. Each dream is mapped to one of the 1000 ImageNet classes. 
We then measure the proportion of dreams assigned to the 100 Mini-ImageNet classes versus the remaining 900 ImageNet classes.

\begin{table*}[ht]
\centering
\caption{\textbf{Relative training cost per task normalized to ER-ACE ($1\times$) on SeqMiniImageNet.} 
ER-ACE serves as the reference cost model. We report training cost under different diffusion usage regimes (none, augmentation/rehearsal with frozen generator, generator-trained replay, and dreaming). 
Inference cost remains identical across methods ($1\times$). A detailed cost decomposition is provided in Appendix~\ref{sec:supp_tflops}.}\label{tab:computation_cost}
\resizebox{\textwidth}{!}{%
\begin{tabular}{l l c c c}
\toprule
\textbf{Method} & \textbf{Diffusion Usage} & \textbf{Training cost} & \textbf{Inference cost} & \textbf{Avg. Acc. ($\uparrow$)} \\
\midrule
ER-ACE & None & $1\times$ & \textbf{$1\times$} & \resultnof{33.26}{3.51} \\
+ Mixup & None & $1\times$ & \textbf{$1\times$} & \resultnof{36.84}{0.94} \\
+ Diffusion (Augmentation + Rehearsal) & Frozen (samples only) & $\sim 7\times$ & \textbf{$1\times$} & \resultnof{36.69}{0.42}  \\
\midrule
DDGR (ResNet) & Generator trained (Replay) & $\sim 489\times$ & \textbf{$1\times$} & \resultnof{43.46}{0.25} \\
\textbf{D2L (Ours)} & Frozen + Dream classes & $\sim 17\times$ & \textbf{$1\times$} & \resultnof{40.90}{0.95} \\
\bottomrule
\end{tabular}
}
\end{table*}

Only 10.11\% of dreamed samples are classified into the 100 Mini-ImageNet classes, while 89.89\% are assigned to the other 900 classes. This distribution closely matches the underlying class prior (10\% vs. 90\%), indicating that dreams are statistically OOD relative to Mini-ImageNet and do not exhibit systematic alignment with target classes.
\subsection{Cost Analysis}
\label{sec:compute_cost}

We evaluate the computational cost of D2L relative to standard rehearsal-based continual learning. ER-ACE is used as the reference cost model and its training cost is normalized to $1\times$. 
All reported values therefore represent relative training cost per task with respect to ER-ACE.

D2L integrates a pre-trained diffusion model while training the discriminative backbone from scratch. The diffusion generator remains frozen and only lightweight soft prompts are optimized. 
For a task with $N$ real classes, D2L introduces $N$ auxiliary dream classes, expanding the label space and increasing the discriminative learning problem.

To isolate whether diffusion alone explains the gains, we construct a controlled baseline where diffusion is used only for data augmentation and rehearsal. For each task with $N$ classes, we generate the same number of synthetic samples as those used in D2L, but we keep the label space unchanged: generated samples are used 
to augment current task classes and to rehearse past classes, without introducing auxiliary dream classes. As shown in Tab.~\ref{tab:computation_cost}, this diffusion-augmented ER-ACE incurs approximately $\sim 7\times$ the cost of ER-ACE yet does not match the performance of D2L (Tab.~\ref{tab:computation_cost}), indicating that the improvements are not due to synthetic sample 
enrichment alone but to the structural effect of auxiliary dream  class generation. 
ER-ACE+Mixup is also reported as a low-cost  reference ($1\times$), but similarly does not reproduce the gains 
of D2L.

Tab.~\ref{tab:computation_cost} also reports performance of DDGR~\cite{yan2023ddgr}, a generative replay method, to avoid conceptual confusion between D2L and diffusion-based generative replay, highlighting the difference between our frozen-generator dreaming process and a generator-trained replay.
Unlike D2L, DDGR requires training or fine-tuning the diffusion generator, which dominates the computational cost and results in an overhead of approximately $\sim 489\times$ relative to ER-ACE, while yielding less than 3 percentage points of additional accuracy.
D2L avoids generator training entirely and requires 
$\sim 17\times$ cost, reflecting additional discriminative optimization due to the expanded label space and prompt  optimization stage. This value represents an upper bound corresponding to early tasks with the maximum number of dream  classes; as fewer auxiliary classes are introduced in later  tasks, the effective overhead decreases (approximately 
$\sim 10\times$ in our estimates). 
Inference cost remains identical ($1\times$) across all methods.
\section{Limitations}

While D2L-generated dreams are qualitatively distant from the target dataset, a residual concern is potential categorical leakage of future-class information. 
To quantify this, we perform a \textit{leak test} using a joint classifier trained on all classes in the sequence.

A \textit{leak} is counted when a future class is assigned to a classifier head that was previously initialized from a dream class and that the joint model 
also maps to that same future class. This captures the scenario in which a dreamed representation unintentionally aligns with a future semantic category.

On Mini-ImageNet, we observe on average 1.76 such leaks, corresponding to 3.93\% of class replacements. These events are therefore rare. Moreover, their occurrence implicitly assumes that the diffusion model can faithfully synthesize future semantic concepts: an assumption not supported by the preceding OOD analysis.

Nonetheless, mitigating this theoretical leakage could involve unlearning future-class concepts from the diffusion model (e.g., via concept erasure or negative-gradient editing) prior to prompt optimization.\\

\section{Conclusion}
\label{sec:conclusion}

In this work, we introduced \textbf{Dream2Learn}, an approach inspired by the ability of the human brain to consolidate past experiences and anticipate future ones through dreaming. 
Our method pairs a classification network with a generative model to synthesize structured training signals, reinforcing past knowledge and enhancing forward transfer---allowing the model to leverage prior knowledge to improve learning on future tasks. 
Experiments on standard continual learning benchmarks show that dreaming helps mitigate forgetting and can support feature learning by expanding the classifier’s representation space,  turning negative forward transfer into positive. Using soft prompt optimization within a latent diffusion model, D2L generates novel yet coherent classes with the oracle model helping to maintain sample quality by preventing collapse.\\
In summary, D2L offers a practical approach to structuring model knowledge over time. By generating intermediate representations, it illustrates the potential of synthetic data to support abstraction and transfer in continual learning. 

\newpage

\begin{appendices}
{
    \makeatletter
    \newcommand{\appsection}[1]{%
      \refstepcounter{section}
      \section*{\appendixname~\thesection. #1}%
    }
    \makeatother
    \clearpage
    \appsection{Metrics for Oracle Input}

\label{sec:oracle_features}
The oracle network takes as input feature sequences $\mathbf{Z}_t = [\zz_{t-k+1}, \dots, \zz_t]$, 
with each $\zz_i \in \mathbb{R}^4$ summarizing properties of the generated samples at iteration $i$. In order to train the oracle network $O$, we initially designed a pool of 25 candidate features capturing different aspects of the generation process. These features can be grouped into three broad categories:
\begin{itemize}
    \item \textbf{Image-level quality metrics}. We compute average SSIM, PSNR, and MSE among pairs of generated images at the same optimization step. These same metrics are also computed between each generated image and its conditioning (target) image. We also compute CLIP-iQA scores (quality, complexity, naturalness, realism) on both generated and target images.
    \item \textbf{Feature-based statistics}. We extract classifier feature-level statistics: cosine similarity and MSE computed either among generated images or between generated and conditioning images. We therefore computed embeddings standard deviation on both generated and target images.
    \item \textbf{Classifier-based uncertainty signals.}  
    From the target classifier logits we compute statistical descriptors including variance, entropy, 
    range (max–min), and kurtosis, averaged across generated samples.  
    We also include the cross-entropy loss signal used during prompt optimization.
\end{itemize}
 
To reduce redundancy and identify the most informative subset, we performed a SHAP-based feature importance analysis across multiple trajectories. Four features consistently ranked highest and were retained for the oracle used in the main experiments.
    \appsection{Computational Cost Estimation}
\label{sec:supp_tflops}

This section describes the formulas and assumptions used to estimate the TFLOP compute budget underlying Tab.~\ref{tab:computation_cost}.
All quantities below are computed \emph{per representative task} (i.e., for a generic task in the stream). For D2L, whenever task-dependent quantities vary, we report the \emph{worst-case} setting used in our accounting.

\subsection{Units and profiling convention}
We estimate compute using a FLOP profiler (THOP in our implementation). In this supplement, we follow the convention
$1\,\mathrm{MAC}=1\,\mathrm{FLOP}$, and we report all compute values directly in TFLOPs by scaling the profiler output by $10^{12}$.
Since our table reports \emph{relative} costs, using a different MAC-to-FLOP convention would not change the conclusions as long as it is applied consistently across methods.

\subsection{Overview of cost terms and normalization (per task)}
We estimate the compute budget \emph{per representative task} by decomposing the total cost into a small set of terms shared across methods:
(i) backbone training (\(\mathcal{C}_{\text{CL}}\)), (ii) optional image synthesis (\(\mathcal{C}_{\text{synth}}\)),
(iii) optional generator training/fine-tuning (\(\mathcal{C}^{\mathcal{G}}_{\text{train}}\)),
and (iv) optional lightweight parameter optimization (e.g., soft prompts; \(\mathcal{C}_{\text{opt}}\)).

Throughout, we denote by \(\mathcal{B}\) the backbone network and by \(\mathcal{G}\) the generator (when present).
Both \(\mathcal{B}\) and \(\mathcal{G}\) may vary depending on the method; consequently, the per-sample forward costs
\(\mathcal{C}^{\mathcal{B}}_{\mathrm{fwd}}\) and \(\mathcal{C}^{\mathcal{G}}_{\mathrm{fwd}}\) are always obtained by profiling the specific backbone/generator
used by each approach.(See cost details in Tab.~\ref{tab:computation_cost_supp}).

Given a representative task, we define the total compute for each method as:
\begin{equation}
\mathcal{C}_{\mathrm{ER\text{-}ACE}} = \mathcal{C}_{\text{CL}},
\end{equation}

\begin{equation}
\mathcal{C}_{\mathrm{DDGR}} = \mathcal{C}_{\text{CL}} + \mathcal{C}^{\mathcal{G}}_{\text{train}} + \mathcal{C}_{\text{synth}},
\end{equation}

\begin{equation}
\mathcal{C}_{\mathrm{D2L}} = \mathcal{C}_{\text{CL}} + \mathcal{C}'_{\text{CL}} + \mathcal{C}_{\text{opt}} + \mathcal{C}_{\text{synth}},
\end{equation}

where \(\mathcal{C}'_{\text{CL}}\) denotes the additional backbone training performed on \emph{dreamed} samples
(~\eqref{eq:cl_dream_cost}).

Finally, Tab.~\ref{tab:computation_cost} reports the \textbf{Relative Training Cost} by normalizing each method to ER-ACE:
\begin{equation}
\mathcal{C}_{\mathrm{rel}}(\text{method})
\;=\;
\frac{\mathcal{C}_{\mathrm{method}}}{\mathcal{C}_{\mathrm{ER\text{-}ACE}}}.
\end{equation}

The following subsections detail the formulas and assumptions used to estimate each term
(\(\mathcal{C}_{\text{CL}}\), \(\mathcal{C}_{\text{synth}}\), \(\mathcal{C}^{\mathcal{G}}_{\text{train}}\), \(\mathcal{C}_{\text{opt}}\)).

\subsection{Backbone training ($\mathcal{C}_{\text{CL}}$)}
Let $N_{\mathrm{img}}$ be the number of real images for the current task, $E$ the number of epochs, and $\rho$ the rehearsal ratio (replay samples per real sample).
We denote by $\mathcal{C}^{\mathcal{B}}_{\mathrm{fwd}}$ the backbone per-sample forward compute.

We approximate one standard training update (forward + backward) via:
\begin{equation}
\kappa_{\mathrm{train}} \triangleq 3.
\end{equation}
Thus, the backbone training compute for the task is
\begin{equation}
\mathcal{C}_{\text{CL}}
\;=\;
N_{\mathrm{img}}
\cdot E
\cdot (1+\rho)
\cdot \kappa_{\mathrm{train}}
\cdot \mathcal{C}^{\mathcal{B}}_{\mathrm{fwd}}.
\label{eq:cl_cost}
\end{equation}

In our accounting we consider seq-MiniImageNet with $5$ classes per task and $500$ images per class (i.e., $N_{\mathrm{img}}=2500$ images per task), trained for $E=10$ epochs. For ER-style training, each minibatch from the current task is paired with a minibatch sampled from the replay buffer, hence we set $\rho=1$.

\paragraph{Diffusion-based Augmentation}
In the case where we use diffusion for data augmentation, we consider an additional factor $\alpha$ to account for the increased number of augmented samples during backbone training. The total backbone training compute $\mathcal{C}_{\text{CL}}$ in this case becomes:
\begin{equation}
\mathcal{C}_{\text{CL}} \;=\;
N_{\mathrm{img}} \cdot (1 +\alpha) \cdot E \cdot (1+\rho) \cdot \kappa_{\mathrm{train}} \cdot \mathcal{C}^{\mathcal{B}}_{\mathrm{fwd}}
\end{equation}
where $\alpha$ represents the augmentation factor introduced by the diffusion process. In our setting $\alpha = 1$.

\paragraph{D2L}
For D2L, the backbone training cost is repeated for both real images and dreamed images. We denote the number of dreamed images as $N_{\mathrm{dream}}$. In our worst-case scenario, where we generate $500$ images per class and consider $45$ dreamed classes, the additional cost is computed as:
\begin{equation}
\mathcal{C}_{\text{CL}}' =  N_{\mathrm{dream}} \cdot E \cdot (1+\rho) \cdot \kappa_{\mathrm{train}} \cdot \mathcal{C}^{\mathcal{B}}_{\mathrm{fwd}},
\label{eq:cl_dream_cost}
\end{equation}
where $N_{\mathrm{dream}} = 500 \cdot 45$ represents the number of dreamed images in the worst case.

\subsection{Image Synthesis ($\mathcal{C}_{\text{synth}}$)}
If a method synthesizes $N_{\mathrm{gen}}$ images, we compute
\begin{equation}
\mathcal{C}_{\text{synth}}
\;=\;
N_{\mathrm{gen}} \cdot \mathcal{C}^{\mathcal{G}}_{\mathrm{fwd}},
\label{eq:synth_cost}
\end{equation}
where $\mathcal{C}^{\mathcal{G}}_{\mathrm{fwd}}$ is the cost to generate \emph{one} image.

\paragraph{D2L}
For D2L, we profile the entire dreaming pipeline end-to-end, including the diffusion, image adapter and continual classifier. Hence, $\mathcal{C}^{\mathcal{G}}_{\mathrm{fwd}}$ for D2L already incorporates \emph{all} components. At inference time, D2L uses LCM with $4$ diffusion steps, and generates a class-balanced set of
\begin{equation}
N_{\mathrm{gen}}^{\mathrm{D2L}} = 500 \cdot K_{\mathrm{dream}},
\end{equation}
where $K_{\mathrm{dream}}$ is the number of dreamed classes. In our accounting of training compute, we use the \textbf{worst-case} setting $K_{\mathrm{dream}} = 45$ (i.e., 45 dreamed classes for the task).

\paragraph{DDGR}
For DDGR, we profile the diffusion generator only. 
In our inference accounting, DDGR uses $4000$ nominal steps with spacing, resulting in $250$ effective denoising evaluations.
DDGR generates a smaller set intended to populate the replay buffer:
\begin{equation}
N_{\mathrm{gen}}^{\mathrm{DDGR}} = 20 \cdot K,
\end{equation}
where $K$ is the number of classes considered for the current task/buffer population. In our setting $K=5$.

\subsection{Generator training ($\mathcal{C}^{\mathcal{G}}_{\text{train}}$)}
DDGR additionally fine-tunes its generator. Let $N_{\mathrm{ft}}$ be the number of generator fine-tuning steps and $\mathcal{C}^{\mathcal{G}}_{\text{fwd}}$ the generator forward cost. We estimate:
\begin{equation}
\mathcal{C}^{\mathcal{G}}_{\text{train}}
\;=\;
N_{\mathrm{ft}}
\cdot \kappa_{\mathrm{train}}
\cdot \mathcal{C}^{\mathcal{G}}_{\text{fwd}},
\qquad
\kappa_{\mathrm{train}}=3.
\label{eq:gen_train_cost}
\end{equation}
In our accounting we use
\begin{equation}
N_{\mathrm{ft}} = 15000 \cdot 64.
\footnote{The DDGR implementation reports $N_{\mathrm{ft}}=15000$ \emph{iterations} with batch size $B=64$; in our accounting this corresponds to $N_{\mathrm{ft}}\!\times\!B$ processed samples. We did not report batch size for other approaches because we fixed $B=1$.}
\end{equation}

\subsection{Soft-prompt optimization ($\mathcal{C}_{\text{opt}}$)}
D2L optimizes a soft prompt while keeping all network weights frozen (diffusion, adapter, classifier).
Therefore, the update cost reduces to forward + backward w.r.t.\ the prompt parameters only, which we approximate as:
\begin{equation}
\kappa_{\mathrm{prompt}} \triangleq 2.
\end{equation}
Let $N_{\mathrm{prompt}}$ be the number of prompts optimized in the task, and let $S_{\mathrm{opt}}$ be the number of optimization steps per prompt. Using the profiled forward cost of the same D2L pipeline, $\mathcal{C}^{\mathcal{G}}_{\text{fwd}}$:
\begin{equation}
\mathcal{C}_{\text{opt}}
\;=\;
N_{\mathrm{prompt}}
\cdot S_{\mathrm{opt}}
\cdot \kappa_{\mathrm{prompt}}
\cdot \mathcal{C}^{\mathcal{G}}_{\text{fwd}}.
\label{eq:prompt_opt_cost}
\end{equation}
In our experiments, we use an average of
\begin{equation}
S_{\mathrm{opt}} = 100
\end{equation}
optimization steps per prompt.

\begin{table*}[t]
    \centering
    \caption{\textbf{Computational Analysis and Paradigm Comparison.} 
    We report the \textbf{Relative Training Cost} normalized to the standard ER-ACE baseline ($1\times$).
    We also report a per-task \textbf{TFLOPs Breakdown} and the corresponding \textbf{Total TFLOPs}.}
    \label{tab:computation_cost_supp}
    \resizebox{\textwidth}{!}{%
    \begin{tabular}{l l l c c}
        \toprule
        \textbf{Method} & \textbf{Train Cost Breakdown} & \textbf{TFLOPs Breakdown} & \textbf{Total TFLOPs} & \textbf{Rel. Train Cost} \\
        \midrule
        ER-ACE & $\mathcal{C}_{\text{CL}}$ 
        & $\mathcal{C}_{\text{CL}}{=}273.61$ 
        & $273.61$ 
        & $1\times$ \\

        \hspace{4pt}$+$  Mixup & $\mathcal{C}_{\text{CL}} + \epsilon$ 
        & $\mathcal{C}_{\text{CL}}{=}273.61,\; \epsilon{\approx}0.0$ 
        & $273.61$ 
        & $1\times$ \\


        \hspace{4pt}$+$ Diffusion & $\mathcal{C}_{\text{CL}} + \mathcal{C}_{\text{synth}}$ 
        & $\mathcal{C}_{\text{CL}}{=}547.21,\; \mathcal{C}_{\text{synth}}{=}1373.21$ 
        & $1920.42$ 
        & $\sim 7\times$ \\

        \midrule

        DDGR (ResNet) & $\mathcal{C}_{\text{CL}} + \mathcal{C}^{\mathcal{G}}_{\text{train}} + \mathcal{C}_{\text{synth}}$ 
        & $\mathcal{C}_{\text{CL}}{=}273.61,\; \mathcal{C}^{\mathcal{G}}_{\text{train}}{=}132473.48,\; \mathcal{C}_{\text{synth}}{=}1149.94$ 
        & $133897.03$ 
        & $\sim489\times$ \\

        \textbf{D2L (Ours)} & $\mathcal{C}_{\text{CL}} + \mathcal{C}_{\text{CL}}' + \mathcal{C}_{\text{opt}} + \mathcal{C}_{\text{synth}}$ 
        & $\mathcal{C}_{\text{CL}}{=}273.61,\; \mathcal{C}_{CL'}{=}2462.45 ,\; \mathcal{C}_{\text{opt}}{=}551.11,\; \mathcal{C}_{\text{synth}}{=}1377.77$ 
        & $4664.93$ 
        & $\sim 17\times$ \\
        \bottomrule
    \end{tabular}
    }
\end{table*}

}
\end{appendices}

\clearpage
\newpage
\bibliographystyle{unsrt}  
\bibliography{bibliography}

\end{document}